\documentclass{article}

\usepackage[final]{corl_2019} 
\usepackage{amsfonts}
\usepackage{amsmath}
\usepackage{graphics}
\usepackage{wrapfig}
\usepackage{subcaption}
\usepackage{float} 
 
\usepackage{epsfig}
\usepackage{lipsum}
\usepackage{todonotes}

\DeclareMathOperator*{\argmax}{arg\,max}

\title{Macro-Action-Based Deep Multi-Agent \\ Reinforcement Learning}

\author{
    Yuchen Xiao \quad \quad
    Joshua Hoffman \quad \quad 
    Christopher Amato\\
    Khoury College of Computer Sciences\\
    Northeastern University,
    United States\\
    \texttt{\{xiao.yuch, hoffman.j\}@husky.neu.edu, c.amato@northeastern.neu}
}

\makeatletter
\let\thetitle\@title
\let\theauthor\@author

\begin{document}
\maketitle


\begin{abstract}
In real-world multi-robot systems, performing high-quality, collaborative behaviors requires robots to asynchronously reason about high-level action selection at varying time durations. Macro-Action Decentralized Partially Observable Markov Decision Processes (MacDec-POMDPs) provide a general framework for asynchronous decision making under uncertainty in fully cooperative multi-agent tasks. However, multi-agent deep reinforcement learning methods have only been developed for (synchronous) primitive-action problems. 
 This paper proposes two Deep Q-Network (DQN) based methods for learning decentralized and centralized macro-action-value functions with novel macro-action trajectory replay buffers introduced for each case. 
 Evaluations on benchmark problems and a larger domain demonstrate the advantage of learning with macro-actions over primitive-actions and the scalability of our approaches.   

\end{abstract}

\keywords{Multi-Agent, Reinforcement Learning, Macro-Actions } 


\section{Introduction}
\vspace{-5pt}
As more robots are deployed in various settings, these robots must be able to act and learn in environments with other agents in them. A number of methods have been developed for solving the resulting multi-robot (or more generally multi-agent) learning problem. In particular, significant progress has been made on multi-agent deep reinforcement learning to solve challenging tasks in cooperative as well as competitive scenarios (e.g.,~\citep{DecHDRQN,foerster:aaai18,lowe2017multi,rashid:icml18}).
However, current methods assume that actions are modeled as primitive operations and synchronized action execution over agents.

In real-world multi-robot cooperative tasks, however, robots often select and complete actions at different times. Such asynchronous collaboration requires a different set of methods that consider these different completion times. Macro-action-based frameworks allow asynchronous action selection and termination while also naturally representing high-level robot controllers (e.g., navigation to a waypoint or grasping an object). 
In the multi-agent case, the 
\textit{Macro-Action Decentralized Partially Observable Markov Decision Process} (MacDec-POMDP)~\citep{AAMAS14AKK,AmatoJAIR19} extends the \textit{options framework}~\citep{Sutton:1999} to partially observable multi-agent domains. 
Planning methods have been developed for MacDec-POMDPs which have been demonstrated in realistic robotics problems~\citep{ICRA15MacDec,RSS15,IJRR17DecPOSMDP,AAAI16}, but only limited learning settings have been considered~\cite{IROS17}.


Nevertheless, a principled way is still missing to generalize the above multi-agent deep reinforcement learning methods to macro-action-based robotics problems. In this paper, we bridge this gap by: (a) proposing a \emph{decentralized} macro-action-based learning method that is based on DQN
~\citep{DQN} and generates Macro-Action Concurrent Experience Replay Trajectories (Mac-CERTs) to properly maintain macro-action trajectories for each agent; (b) introducing a \emph{centralized} macro-action-based learning method that is also based on DQN and generates  Macro-Action Joint Experience Replay Trajectories (Mac-JERTs) to maintain time information in macro-action trajectories along with a conditional target prediction method for learning a centralized joint macro-action-value function. 
Decentralized learning of decentralized policies is needed for online learning by the agents, but is difficult due to the noisy and limited learning signals of each agent and the apparent non-stationarity of the domain. Centralized learning of centralized policies is important when full communication is available during execution or as an intermediate step in generating decentralized policies in a centralized manner. 
To our knowledge, this is the first formalization of macro-action-based multi-agent deep reinforcement learning under partial observability.

We test our methods in simulation against state-of-the-art (primitive-action) methods. 
The results demonstrate that our methods are able to achieve much higher performance than learning with primitive actions and are scalable to large environment spaces. 
We believe these methods are promising for learning in realistic multi-robot settings. 








\section{Background}
\label{sec:background}
\vspace{-5pt}

We develop decentralized and centralized learning methods for decentralized and centralized execution, respectively, using the MacDec-POMDP framework. We first describe the Dec-POMDP models and deep RL methods that our approaches build upon. 

\subsection{Dec-POMDPs and MacDec-POMDPs}

We focus on fully cooperative decentralized multi-agent domains with both state and outcome uncertainties. As such, each agent must choose actions individually 
 purely based on local observations. This setting is described as a decentralized partially observable Markov decision process (Dec-POMDP)~\citep{Oliehoek}, 
formally represented as a tuple $\langle I, S, A, \Omega, T, O, R \rangle$, where $I$ is a finite set of agents; $S$ is a finite set of environment states; $A=\times_iA_i$ is the set of joint actions with $A_i$ being the available actions for each agent $i$; $\Omega=\times_i\Omega_i$ is the set of joint observations with $\Omega_i$ being the set of observations for each agent $i$; 
At each time-step, the environment state transits from $s$, after taking a joint action $\vec{a}$, to a new state $s'$ according to the state transition function $T(s, \vec{a}, s') = P(s'\mid s,\vec{a})$. $O(\vec{o}, \vec{a}, s')=P(\vec{o}\mid\vec{a}, s')$ denotes the probability of receiving a joint observation $\vec{o}$ when a joint action $\vec{a}$ were taken and arriving in state $s'$. $R:S\times A\rightarrow \mathbb{R}$ is a reward function assigning a shared immediate reward $r(s,\vec{a})$ for taking $\vec{a}$ in $s$. Due to the partial observability, the policy $\pi_i$ maintained by each agent $i$ is a mapping from local observation histories to actions. In finite horizon Dec-POMDPs, the objective of solution methods is to find a joint policy $\pi=\times_i\pi_i$ that maximizes the expected sum of discounted rewards starting from $s_0$, $V^{\pi}(s_{(0)})=\mathbb{E}\big[\sum_{t=0}^{\mathbb{H}-1}\gamma^tr(s_{(t)},\vec{a}_{(t)})\mid s_{(0)}, \pi\big]$, where $\gamma\in[0,1]$ is a discount factor, and $\mathbb{H}$ is the horizon of the problem.

Dec-POMDPs with temporally extended actions that are based on the option framework~\citep{Sutton:1999}  are referred to MacDec-POMDPs~\citep{AAMAS14AKK,AmatoJAIR19}.
 Formally, a MacDec-POMDP is represented as a tuple $\langle I, S, A, M, \Omega, \zeta, T, O, Z, R\rangle$, where $I$, $S$, $A$, $\Omega$, $O$, $R$ are the same as defined in Dec-POMDP; $M=\times_iM_i$ is the set of joint macro-actions with $M_i$ being a finite set of macro-actions for each agent $i$; $\zeta=\times_i\zeta_i$ is the set of joint macro-observations with $\zeta_i$ being a finite macro-observation space for each agent $i$. Each macro-action is defined as a tuple $m = \langle\beta_{m}, I_{m}, \pi_{m} \rangle$, where the stochastic termination condition $\beta_{m}:H_i^A\rightarrow[0,1]$ and the initiation set $I_{m}\subset H^{M}_i$ of the corresponding macro-action $m$, respectively, depend on agent $i$'s primitive-action-observation history $H_i^A$ and macro-action-observation history $H_i^M$; $\pi_{m}:H_i^A\rightarrow A_i$, denotes the low-level policy to achieve the macro-action $m$. Taking into account the stochastic termination of a macro-action, the transition probability is rewritten as  $T(s',\vec{\tau}, s, \vec{m}) = P(s',\vec{\tau}\mid s, \vec{m})$, where $\vec{\tau}$ is the number of time-steps taken by the joint macro-action $\vec{m}$ that terminates when \emph{any agent} completes its own macro-action;
$Z(\vec{z}, \vec{m},s') = P(\vec{z}\mid \vec{m},s')$ denotes the joint macro-observation likelihood model. The objective is then to find a joint high-level policy that chooses only at the macro-action level $\Psi = \times_i \Psi_i$ such that the value of $\Psi$ from the initial state $s_0$, $V^{\Psi}(s_{(0)})=\mathbb{E}\big[\sum_{t=0}^{\mathbb{H}-1}\gamma^tr(s_{(t)},\vec{a}_{(t)})\mid s_{(0)}, \pi, \Psi\big]$ is optimized.

\subsection{Deep Q-Networks and Deep Recurrent Q-Networks}

Q-learning~\citep{Watkins1992} is a popular model-free method to optimize a policy $\pi$ by iteratively updating an action-value function $Q(s,a)$.
Deep Q-networks (DQN) \citep{DQN} extend Q-learning to include a deep neural net as a function approximator. DQN learns $Q_\theta(s,a)$, parameterized with $\theta$, by minimizing the loss:
$\mathcal{L}(\theta)=\mathbb{E}_{<s, a, s' r>\sim\mathcal{D}}\Big[\big(y - Q_{\theta}(s, a)\big)^2 \Big] \text{, where\,\, } y=r + \gamma\argmax_{a'}Q_{\theta^-}(s',a')$. A target action-value function $Q_{\theta^-}$ and an experience replay buffer $\mathcal{D}$~\citep{Lin1992} are implemented for stable learning. In order to deal with the maximum bias, the idea behind Double Q-learning~\citep{DoubleQ} is generalized to DQN, called Double DQN, by rewriting the target value calculation as $y = r + \gamma Q_{\theta^-}(s', \argmax_{a'}Q_{\theta}(s',a'))$~\citep{DDQN}. Deep Recurrent Q-Networks (DRQN) is proposed to handle single agent tasks with partial observability~\citep{DRQN}, where a recurrent layer (LSTM~\citep{LSTM}) is applied to maintain an internal hidden state which is referred to as the history. In our work, we extend Double DQN with a recurrent layer, called DDRQN, to learn macro-action-based policies. This is done for decentralized policies in Section~\ref{macdec} and centralized policies in Section~\ref{maccen}. 

\subsection{Decentralized Hysteretic Deep Recurrent Q-Networks}

Many methods have extended Deep Q-learning to Dec-POMDPs (e.g.,~\citep{DecHDRQN,foerster:aaai18,rashid:icml18}). One of these methods, called Dec-HDRQN~\citep{DecHDRQN}, is a decentralized learning method that generalizes Hysteretic Q-learning~\citep{HQL}, which uses two learning rate $\alpha$ and $\beta$ to update the action-value function, to DQN and DRQN. 
Specifically, $\alpha$ is a normal learning rate used when TD error is positive, and $\beta$ is a smaller learning rate used otherwise. This facilitates multi-agent learning by making each agent robust against negative updating due to teammates' mistakes. Negative TD error is assumed to be due to other agent exploration rather than domain stochasticity thereby avoiding convergence to local optima in some domains. 
Decentralized learning is particularly difficult because the environment seems non-stationary from each single agent's perspective (i.e., all the other agents are considered to be part of the environment, but they are also learning and exploring). 
A new replay buffer called Concurrent Experience Replay Trajectories (CERTs) 
is also implemented with Dec-HDRQN to assist with the non-stationarity issue, by sampling concurrent experiences for training, which encourages each agent's policy to be optimized toward same direction. In this paper, we propose an extension of CERTs (Section~\ref{macdec}) such that Dec-HDRQN is able to learn macro-action-based policies.

	





\section{Approach}
\label{sec:approach}
\vspace{-5pt}

In multi-robot deep reinforcement learning with macro-actions, the highly asynchronous execution of macro-actions motivates a need for a principled way for updating values and maintaining replay buffers. 
In this section, we introduce two approaches for solving these problems 
for learning decentralized (Section \ref{macdec}) and centralized (Section \ref{maccen}) policies. In each case, we assume the agent(s) can observe the current macro-action, macro-observation and reward at each time-step. That is, we do not have access to the primitive-level actions and observations, but we could indirectly calculate the duration of a macro-action by counting time-steps. 

\subsection{Learning Decentralized Policy with Macro-Actions}
\label{macdec}

\begin{figure}[h!]
    \centering 
    \vspace{-7mm}
    \includegraphics[height=3.6cm]{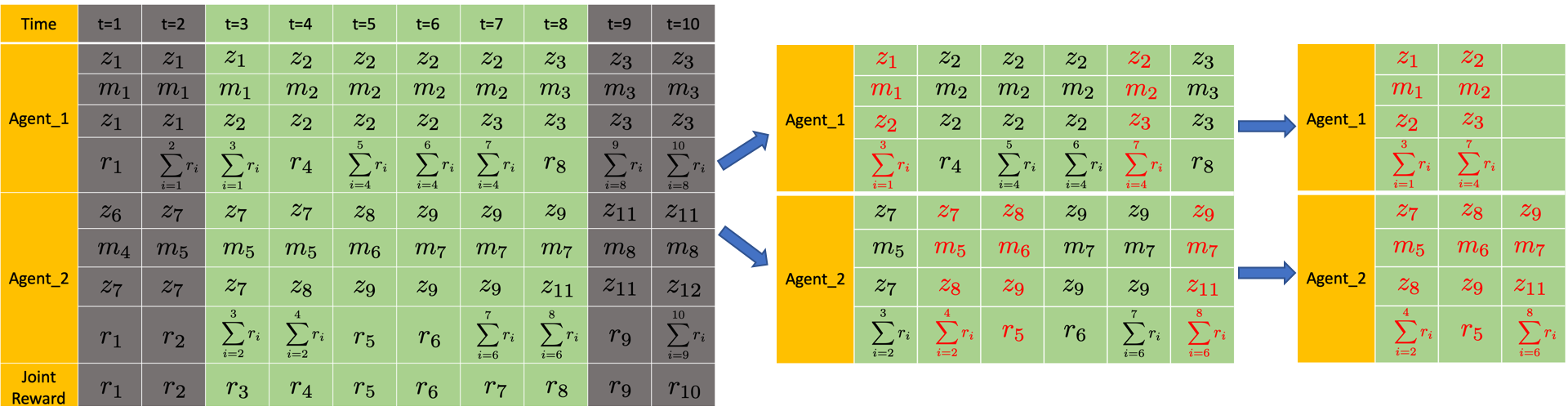}
    \vspace{-1mm}
    \caption{A example of Mac-CERTs. Two agents first sample concurrent sequential experiences (green area) from the replay buffer; The valid experience (when the macro-action terminates, marked as red), are then selected out to compose a squeezed sequential experiences for each agent. }
    \label{decbuffer}
    \vspace{-4mm}
\end{figure}

In the decentralized case, each agent only has access to its own macro-actions and macro-observations as well as the joint reward at each time-step. 
As a result, there are several choices for how information is maintained. For example, each agent could maintain exact the information mentioned above (as seen on the left side of Fig.~\ref{decbuffer}), the time-step information can be removed (losing the duration information), or some other representation could be used that explicitly calculates time. We choose the middle approach. As a result, updates only need to take place for each agent after the completion of its own macro-action, and we introduce a replay buffer based on Macro-Action Concurrent Experience Reply Trajectories (Mac-CERTs) visualized in Fig.~\ref{decbuffer}.


In particular, under a macro-action-observation history $h$, 
each agent independently selects a macro-action $m$ and maintains an accumulating reward, 
$r^c(h,m,\tau) = \sum_{t=t_{m}}^{t_{m}+\tau-1}\gamma^{t-t_{m}} r_t$
for the macro-action from its first time-step $t_{m}$ to the termination step $t_{m}+\tau-1$. 
The agent then obtains a new macro-observation $z'$ with the probability $P(z'\mid h,m,\tau)$ and results in a new history $h'=\langle h,m,z'\rangle$ under the transition model $P(h',\tau\mid h,m)$. 
Accordingly, the experience tuple collected by each agent $i$ is represented as $\langle z, m, z', r^c  \rangle_i$, where $z$ is the macro-observation used for choosing the macro-action $m$. 
Note that, if $m$ is still being executed, $z'$ is set to be the same as $z$ (shown in Fig.\ref{decbuffer}). We can write the Bellman equation for each agent $i$ under a given high-level policy $\Psi_i$ as :
\\[-6pt]
\begin{equation}
    Q^{\Psi_i}(h,m) = \mathbb{E}_{\tau \sim \beta_m(h)}\Big[r^c(h,m,\tau) + \gamma^{\tau}\sum_{z'\in\zeta_i}P(z' \mid h, m, \tau)V^{\Psi_i}(h')\Big] \\
\label{bellman1}
\end{equation}
\\[-12pt]

In each training iteration, agents first sample a concurrent mini-batch of sequential experiences (either random traces with same length or entire episodes) from the replay buffer $\mathcal D$. Each sampled sequential experience is further cleaned up by filtering out the experience when the corresponding macro-action is still running. This disposal procedure results in a mini-batch of `squeezed' sequential experiences for each agent's training. A specific example is shown in Fig.~\ref{decbuffer} (assuming $\gamma=1$).       

In this paper, we implement Dec-HDRQN with Double Q-learning (Dec-HDDRQN) to train the decentralized macro-action-value function $Q_{\theta_i}(h, m)$ (in Eq.~\ref{bellman1}) for each agent $i$. Each agent updates its own macro-action-value function by minimizing the loss:
\\[-14pt]
\begin{equation}
    \mathcal{L}(\theta_i)=\mathbb{E}_{<z, m, z', r^c>_i\sim\mathcal{D}}\Big[\bigl(y_i - Q_{\theta _i}(h, m)\bigr)^2 \Big] \text{, where } y_i = r^c + \gamma^\tau Q_{\theta_i^-}\bigl(h', \argmax_{m'} Q_{\theta_i}(h',m')\bigr)
\end{equation} 
\\[-11pt]

\subsection{Learning Centralized Policy with Macro-Actions}
\label{maccen}

Achieving centralized control in the macro-action setting needs to learn a joint macro-action-value function $Q(\vec h, \vec m)$. This requires a way to correctly accumulate the rewards for each joint macro-action.  This is actually more complicated than the decentralized case because there is no obvious update step (i.e., there may never be a time when all agents have terminated their macro-actions at the same time). As a result, we use the idea of updating when \emph{any} agent terminates a macro-action \cite{AAMAS14AKK,AmatoJAIR19}. But this makes updating and maintaining a buffer more complicated than in Section \ref{macdec}. 

\begin{figure}[t!]
    \vspace{-1mm}
    \centering
    \includegraphics[height=2.8cm]{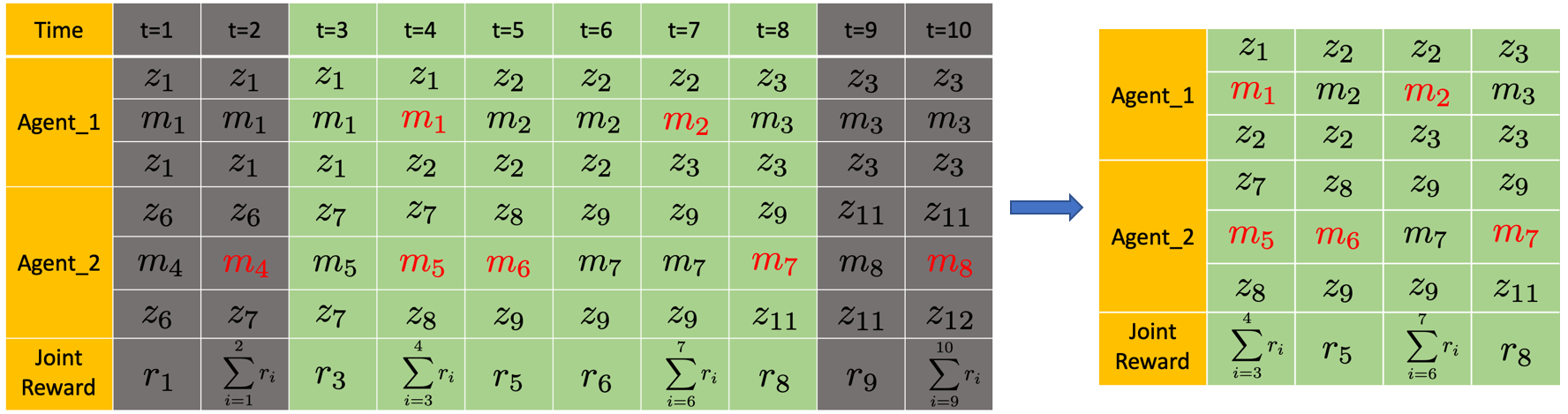}
    \vspace{-1mm}
    \caption{An example of Mac-JERTs. A joint sequential experiences (green area) is first sample from the memory buffer, and then, depending on the termination of each joint macro-action, a squeezed sequential experiences is generated for the centralized training. Each agent's macro-action, which is responsible for the termination of the joint one, is marked in red.}
    \vspace{-5mm}
    \label{cenbuffer}
\end{figure}

In this section, we introduce a centralized replay buffer that we call Macro-Action Joint Experience Replay Trajectories (Mac-JERTs). Instead of independently accumulating the rewards for the corresponding macro-action, in Mac-JERTs, agents share a joint cumulative reward 
$\vec{r}^{\,c}(\vec{h}, \vec{m}, \vec{\tau}) = \sum_{t=t_{\vec{m}}}^{t_{\vec{m}}+\vec{\tau}-1}\gamma^{t-t_{\vec{m}}} r_t$,
where $t_{\vec{m}}$ is the time-step when the joint macro-action $\vec{m}$ starts, 
and $t_{\vec{m}}+\vec{\tau}-1$ is the ending time-step of $\vec{m}$ when \emph{any} agent finishes its macro-action.
For example, in Fig.~\ref{cenbuffer}, the first joint macro-action of the two agents is $\langle m_1, m_4\rangle$ with a length of two time-steps (because Agent\_$2$ accomplished $m_4$ at the second time-step, the joint macro-action then became $\langle m_1,m_5 \rangle$ at the next. As a result, the corresponding cumulative rewards is $\vec{r}^{\,c}=r_1+r_2$, assuming $\gamma=1$. 

In the execution phase, at every step a joint experience, represented as a tuple $\langle\vec{z}, \vec{m}, \vec{z}\,', \vec{r}^{\,c}\rangle$, is collected into the Mac-JERTs. Here, we can write down the Bellman equation under a joint macro-action policy $\Psi$~\citep{AAMAS14AKK}:  
\\[-10pt]
\begin{equation}
    Q^{\Psi}(\vec{h},\vec{m}) = \mathbb{E}_{\vec{\tau}\sim \beta_{\vec{m}}(\vec{h})}\Bigr[\vec{r}^{\,c}(\vec{h},\vec{m},\vec{\tau}) + \gamma^{\vec{\tau}}\sum_{\vec{z}\,'\in\zeta}P(\vec{z}\,' \mid \vec{h}, \vec{m}, \vec{\tau})V^{\Psi}(\vec{h}')\Bigr] 
\label{bellman2}
\end{equation}
\\[-18pt]

In our work, we use Double-DRQN (DDRQN) to train the centralized macro-action-value function. In each training iteration, a mini-batch of sequential joint experiences is first sampled from Mac-JERTs, and then a similar filtering operation, as presented in Section~\ref{macdec}, is used to obtain the `squeezed' joint experiences (shown in Fig.~\ref{cenbuffer}). But, in this case, only one joint reward is maintained that accumulates from the selection of any agent's macro-action to the completion of any (possibly other) agent's macro-action. 

Using the squeezed joint sequential experiences, the centralized macro-action-value function (in Eq.~\ref{bellman2}) at time-step $t$, $Q_{\phi}(\vec{h}_{(t)}, \vec{m}_{(t)})$, is trained end-to-end to minimize the following loss:
\\[-9pt]
\begin{equation}
    \mathcal{L}(\phi)=\mathbb{E}_{<\vec{z}_{(t)}, \vec{m}_{(t)}, \vec{z}_{(t+1)}, \vec{r}^c_{(t)} >\sim\mathcal{D}}\biggr[\Bigr(y_{(t)} - Q_{\phi}\bigr(\vec{h}_{(t)}, \vec{m}_{(t)}\bigr)\Bigr)^2 \biggr] \text{, where}
\end{equation}
\\[-13pt]
\begin{equation}
    \label{uncondi}
    y_{(t)} = \vec{r}^{\,c}_{(t)} + \gamma^{\vec{\tau}} Q_{\phi-}\Bigr(\langle\vec{h}_{(t)}, \vec{z}_{(t+1)}\rangle,\argmax_{\vec{m}'}Q_{\phi}\bigr(\langle\vec{h}_{(t)},\vec{z}_{(t+1)}\rangle,\vec{m}'\bigr)\Bigr)
\end{equation}
\\[-6pt]
The next joint macro-action selection part in Eq.~\ref{uncondi} implies that at the next step all agents will switch to a new macro-action. However, this is often not true. For example, in Fig.~\ref{cenbuffer}, the last three squeezed sequential experiences show that only one of the agents changes its macro-action per step. Therefore, the more agents that are not switching macro-actions, the less accurate the prediction that Eq.~\ref{uncondi} will make. In order to have a more correct value estimation for a joint macro-action, here, we propose a \emph{conditional target prediction} as:
\\[-10pt]
\begin{equation}
    \label{condi}
    y_{(t)} = \vec{r}^{\,c}_{(t)} + \gamma^{\vec{\tau}} Q_{\phi-}\Bigr(\langle\vec{h}_{(t)},\vec{z}_{(t+1)}\rangle, \argmax_{\vec{m}'}Q_{\phi}\bigr(\langle\vec{h}_{(t)}, \vec{z}_{(t+1)}\rangle, \vec{m}' \mid \vec{m}^{\text{undone}}_{(t)}\bigr)\Bigr)
\end{equation}
\\[-8pt]
where, $\vec{m}^{\text{undone}}_{(t)}$ is the joint-macro-action over the agents who have not terminated the macro-actions at time-step $t$ and will continue running next step. The comparison of the training results using the two different predictions is discussed in Section~\ref{LCP}.


\section{Experimental Settings}
\label{ES}
\begin{figure}[h!]
    \vspace{-2mm}
    \centering
    \begin{subfigure}{.3\textwidth}
        \centering
        \includegraphics[height=2.4cm]{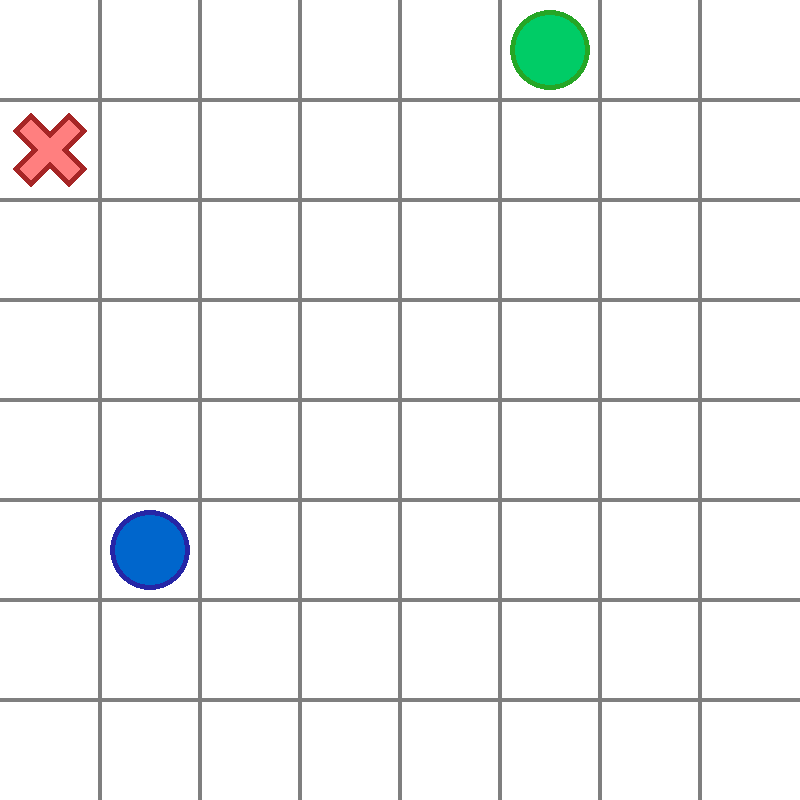}
        \vspace{-1mm}
        \caption{Capture Target}
        \label{ctma}
    \end{subfigure}
    \begin{subfigure}{.3\textwidth}
        \centering
        \includegraphics[height=2.4cm]{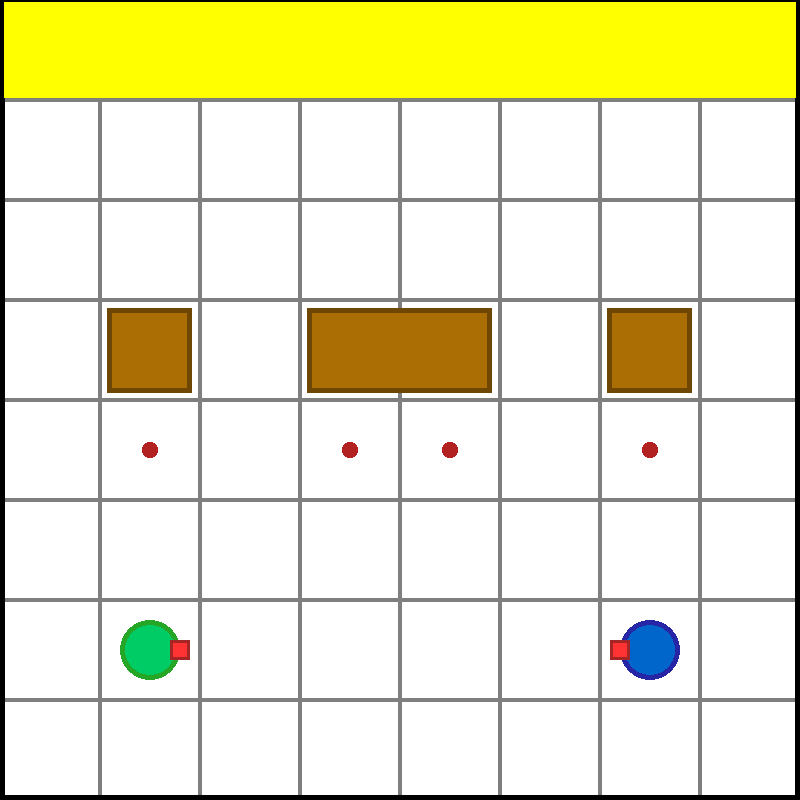}
        \vspace{-1mm}
        \caption{Box Pushing}
        \label{bpma}
    \end{subfigure}
    \begin{subfigure}{.3\textwidth}
        \centering
        \includegraphics[height=2.4cm]{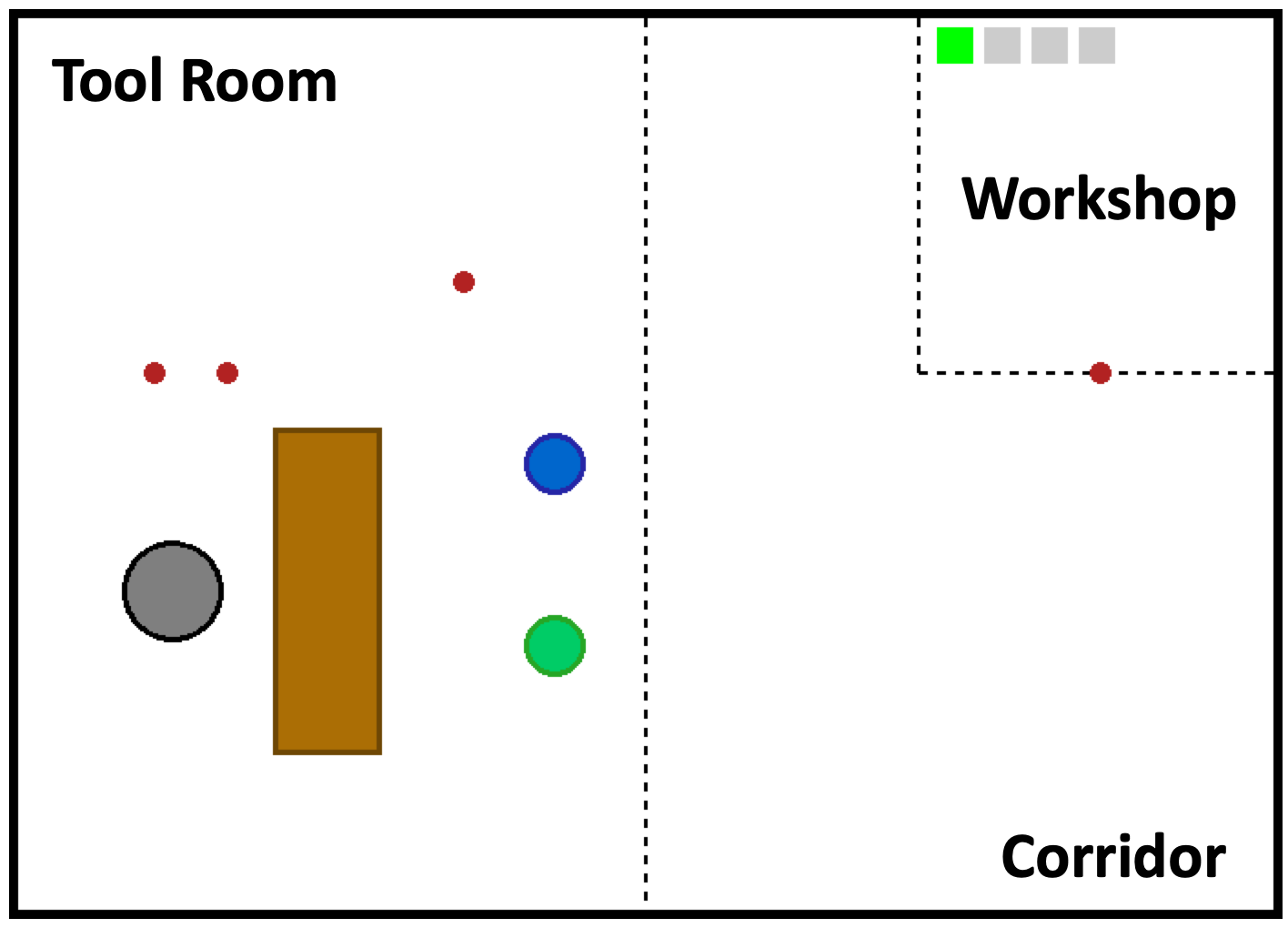}
        \vspace{-1mm}
        \caption{Warehouse Tool Delivery}
        \label{wtdma}
    \end{subfigure}
    \vspace{-1mm}
    \caption{Experimental environments}
    \label{domains}
    \vspace{-5mm}
\end{figure}

\vspace{-5pt}
We evaluate our approaches on three different domains (Fig.~\ref{domains}): (a) Capture Target, a variant of an existing multi-agent-single-target (MAST) domain~\citep{DecHDRQN}; (b) Box Pushing, a benchmark Dec-POMDP domain~\citep{SZuai07}; 
(c) Warehouse Tool Delivery Domain inspired by human-robot interaction. 
Note that the macro-actions, defined in domains that we consider in this paper, are quite simple. It will not always be so straightforward, but we leave macro-action design and selection for future work. 

\subsection{Capture Target with Macro-Actions (CTMA)}

In Fig.~\ref{ctma}, two robots (green and blue circles) are tasked with capturing a randomly moving target (red cross). A terminal reward +1  can only be obtained when the two robots capture the target simultaneously. The macro-observations here are the same as the primitive (low-level) ones: each agent's own location (fully observable) and the target's location (partially observable with a flickering probability of 0.3). In the primitive action version~\citep{DecHDRQN}, each agent has four moving actions (\emph{up, down, left, right}) and a \emph{stay} action. In the macro-action case, there are only two macro-actions for each agent: \textbf{\textit{Move\_to\_Target}}, navigates the robot towards the target and keeps updating the target's location according to the low-level observation; It terminates when the robot reaches the observed target's position. Note that if the target is flicked, agent will continue moving towards the previously observed one; \textbf{\textit{Stay}}, is same as the primitive one and lasts only 1 time-step.     

\vspace{-1mm}
\subsection{Box Pushing with Macro-Actions (BPMA)}

This is a well-known cooperative robotics problem presented in~\citep{SZuai07}. Fig.~\ref{bpma} displays one example of this problem in a grid world. Here, there are two small boxes and one big box in the environment. The goal of the two robots is to cooperatively push the big box, which cannot be moved by each agent on its own, to the yellow area for a higher credit than individually pushing a small box.       

In the primitive action version, each agent has four actions: \textit{move forward}, \textit{turn left}, \textit{turn right} and \textit{stay}. The small box moves forward one grid cell when any robot faces it and executes the \textit{move forward} action. The big box is only movable when the two robots face it in two parallel cells and move forward together. The robot can only observe one of five states in the cell in front of it: empty, teammate, boundary, small box, or big box. During execution, the agents get $-0.1$ reward per step. Successfully pushing the big box to the goal area results in a $+100$ reward or a $+10$ reward for each small box. Either hitting the boundary or pushing the big box alone generates a $-5$ penalty.

In the macro-action version, besides the one-step macro-actions \textbf{\textit{Turn\_left}}, \textbf{\textit{Turn\_right}}, and \textbf{\textit{Stay}}, we include three long-term macro-actions: \textbf{\textit{Move\_to\_small\_box$(i)$}}, navigates the robot to the red waypoint below one of the small boxes and ends with facing the box; 
\textbf{\textit{Move\_to\_big\_box(i)}}, navigates the robot to one of the waypoints below the big box and facing it; \textbf{\textit{Push}}, lets the robot keep moving forward until touching the environment's boundary, hitting the big box on its own, or pushing a box to the goal area.  
Note that, the boxes are only allowed to be pushed toward north, and each episode terminates either one of the boxes pushed to the goal area or after a certain amount of time-steps.    

\subsection{Warehouse Tool Delivery with Macro-Actions (WTDMA)}
\label{WTDD}

\textbf{Task Specification.} In order to test if our approach is scalable to a larger domain requiring more complicated collaborations and long-term reasoning, we designed this Warehouse Tool Delivery problem (Fig.~\ref{wtdma}). 
This environment is a $5\times7$ continuous space, which involves one human working on an assembling task in the workshop. 
The progress bar on the top indicates the total number of steps in the task, the current step (green) the human is working on, and the completed step (black). 
The human always starts from step one and needs a particular tool for each future step to continue. A Fetch robot (gray circle), mounted with a manipulator placed in the tool room, is responsible for searching for the correct tool on the tabletop (brown) and passing it to one of the mobile Turtlebots (green and blue circles) to complete the delivery to the human in time. 
In our experiments, the assembling task has $4$ steps in total, and the time cost on each is $18$. 
Note that, the human is only allowed to get the tool for the next one step from Turtlebots. 
Each episode ends after $H=150$ time-steps, or the human obtains the tool for the last step.

\textbf{Macro-Actions.} Three macro-actions are available for each Turtlebot: \textbf{\textit{Go\_to\_WS}}, navigates the robot to the red waypoint at the workshop, and the length of this action depends on the robot's moving speed $v$ (0.6 in our case); \textbf{\textit{Go\_to\_TR}}, directs the robot to the waypoint located at upper right of the tool room; \textbf{\textit{Get\_Tool}}, leads the robot to the pre-allocated waypoint beside the table and wait there. This action will not terminate until either obtaining one tool from the Fetch robot or after 10 time-steps have passed. There are four macro-actions for the Fetch robot: \textbf{\textit{Wait\_T}} (1 step cost), waits for Turtlebots; \textbf{\textit{Search\_Tool$(i)$}} (6 steps cost), searches for a tool $i$ and place it at a waiting spot on the table; \textbf{\textit{Pass\_to\_T$(i)$}} (4 steps cost), passes one of found tools to Turtlebots $i$. 
Note that: there are only two available waiting spots on the table. If they are both occupied and Fetch still executes \textbf{\textit{Search\_Tool}}, it will be frozen for 6 time-steps. Tools are passed in the order as they found. 

\textbf{Macro-Observations.} Turtlebot can capture four different features in one macro-observation: its own location, the current step the human is working on (only observable in the workshop), the tools being carried by itself, and the number of the tools at the waiting spots (only observable in the tool room). 
Fetch is allowed to observe the number of tools waiting to be passed to Turtlebots, and which Turtlebot is beside the table. Importantly, neither Fetch nor Turtlebot has the knowledge about the correct tool the human needs each step, such that the robots have to reason about this via training. 

\textbf{Rewards.} In order to encourage the robots to deliver the object(s) as soon as possible and to avoid making the human wait, a negative reward $-1$ is issued each time-step. Successfully delivering the correct tool to the human results in a reward $+100$. Additionally, a penalty -10 is allocated to the team when Fetch executes \textbf{\textit{Pass\_to\_T$(i)$}} but no any Turtlebot beside the table.      

\section{Results}
\label{Result}
\vspace{-5pt}

In this section, the performance of our approach on learning decentralized policies in the capture target and box pushing domains are first presented (Section~\ref{LDP}). Then, we show the evaluations on learning centralized policies in the box pushing domain, and also compare 
 training via \textit{conditional target prediction} (Eq.~\ref{condi}) and the \textit{unconditional one} (Eq.~\ref{uncondi}) (Section~\ref{LCP}). Finally, we demonstrate (as expected) that our centralized learning approach enables the robots to learn complex collaborative behaviors in the warehouse domain (Section~\ref{WTD}). 
The results shown below (Fig.~\ref{cp_dec} - Fig.~\ref{result_wtd}) are the mean of the episodic evaluation discounted returns (evaluation performed every 10 training episodes) over 40 runs with the standard error, and further smoothed by averaging over 10 neighbors. Optimal returns are shown as dash-dot lines. Readers are referred to the supplement for the full results.

\subsection{Comparison on Learning Decentralized Policies}
\label{LDP}

We first compare our decentralized approaches in the capture target and box pushing domains.
The experiments in target domain use two MLP layers (32 neurons on each), one LSTM layer~\citep{LSTM} (64 hidden units), and another two MLP layers (32 neurons on each), which is the same architecture as seen in~\citep{DecHDRQN} except using a Leaky Relu layer instead of the regular Relu one as the activation function. In box pushing domain, we tune the number of the neurons in the LSTM layer down to 32. 
\begin{figure}[t]
    \vspace{-2mm}
    \centering
    \begin{subfigure}{.45\textwidth}
        \centering
        \captionsetup{justification=centering}
        \includegraphics[height=3.2cm]{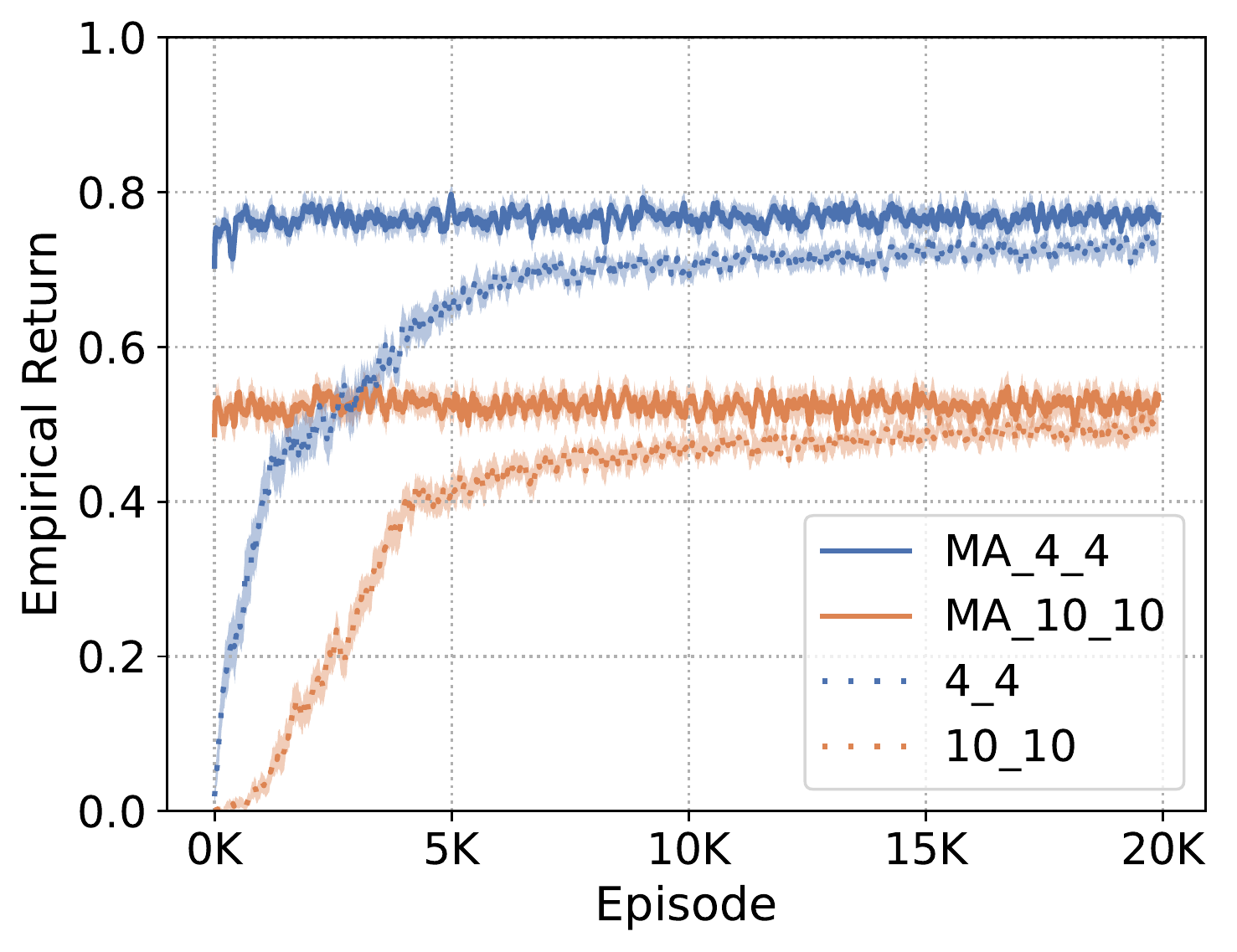}
        \vspace{-2mm}
        \caption{Capture target domains with two grid world sizes: $4\times4$ and $10\times10$}
        \vspace{-1mm}
        \label{ctma_vs_ct}
    \end{subfigure}
    \qquad
    \begin{subfigure}{.45\textwidth}
        \centering
        \captionsetup{justification=centering}
        \includegraphics[height=3.2cm]{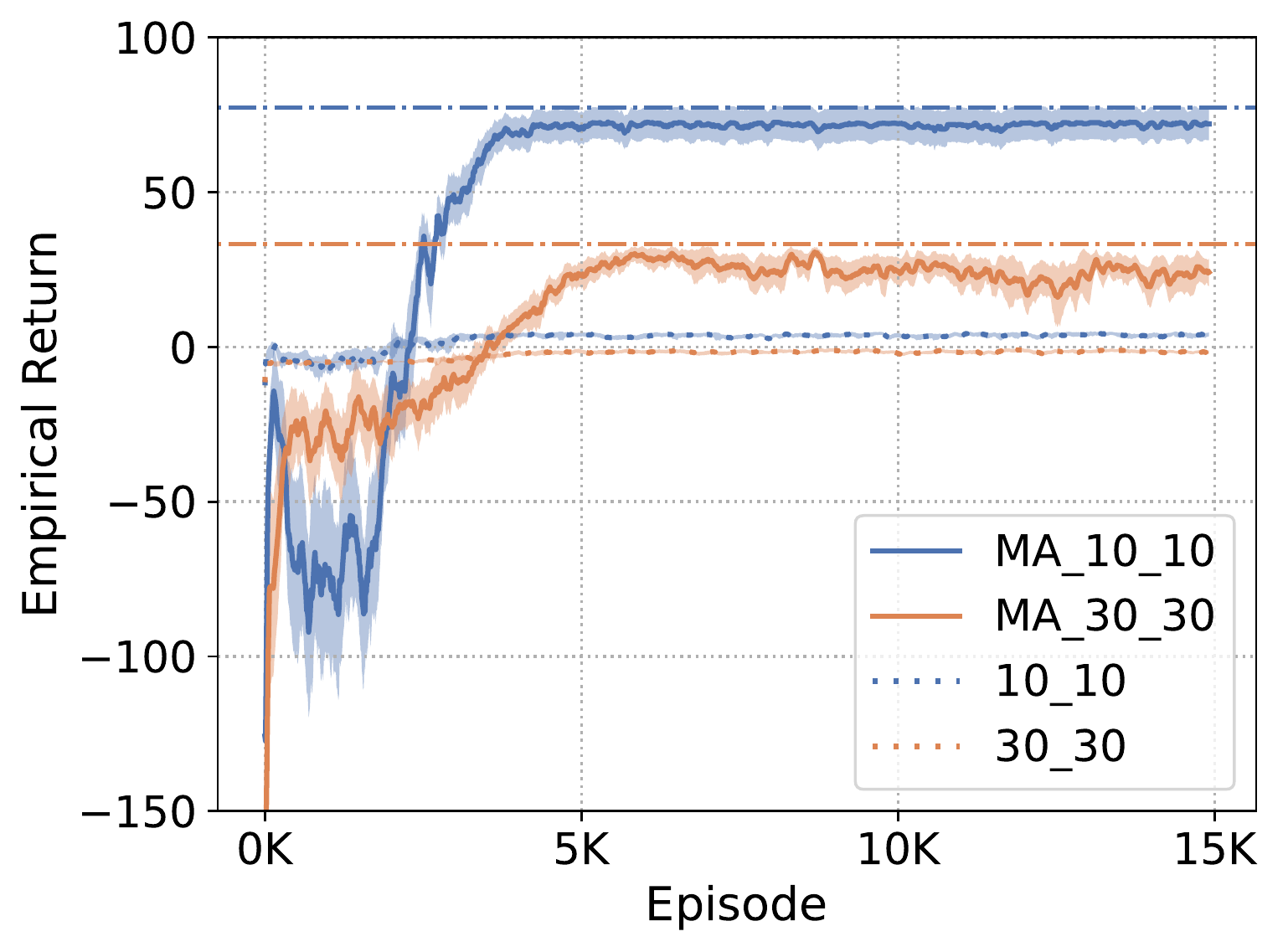}
        \vspace{-2mm}
        \caption{Box pushing domains with two grid world sizes: $10\times10$ and $30\times30$}
        \vspace{-1mm}
        \label{bpma_vs_bp}
    \end{subfigure}
    \caption{Learning decentralized policy with macro-actions (MA) versus primitive actions in capture target ($\gamma=0.95$) and box pushing ($\gamma=0.98$) domains.}
    \vspace{-2mm}
    \label{cp_dec}
\end{figure}

In capture target domain, the macro-actions design provides a smaller action space than the primitive version, which makes the problem easier, and facilitates the agents to learn the good policies much faster to reach the returns that the primitive learner takes longer time to converge towards (Fig.~\ref{ctma_vs_ct}). In box pushing domain, learning with macro-actions achieves near-optimal performance (Fig.~\ref{bpma_vs_bp}), such that two agents behave cooperation to push the big box, rather than pushing the small one on each own learnt under primitive actions setting. Also, near-optimal performance can always be achieved by the macro-actions learner even when the world space increases (e.g. $30\times30$), but the primitive-actions learner cannot.  



\subsection{Results on Learning Centralized Macro-Action Policies}
\label{LCP}

Our approach on learning centralized macro-action based policy is evaluated in box pushing domains. 
The centralized policies are parameterized by the same network architecture in Section~\ref{LDP}. Particularly, 32 neurons in each MLP layers and 64 neurons in LSTM are used for the grid world size smaller than $10 \times 10$, otherwise 64 neurons in each MLP layers.  


\begin{figure}[t]
    \vspace{-1mm}
    \centering
    \begin{subfigure}{.47\textwidth}
        \centering
        \captionsetup{justification=centering}
        \includegraphics[height=3.2cm]{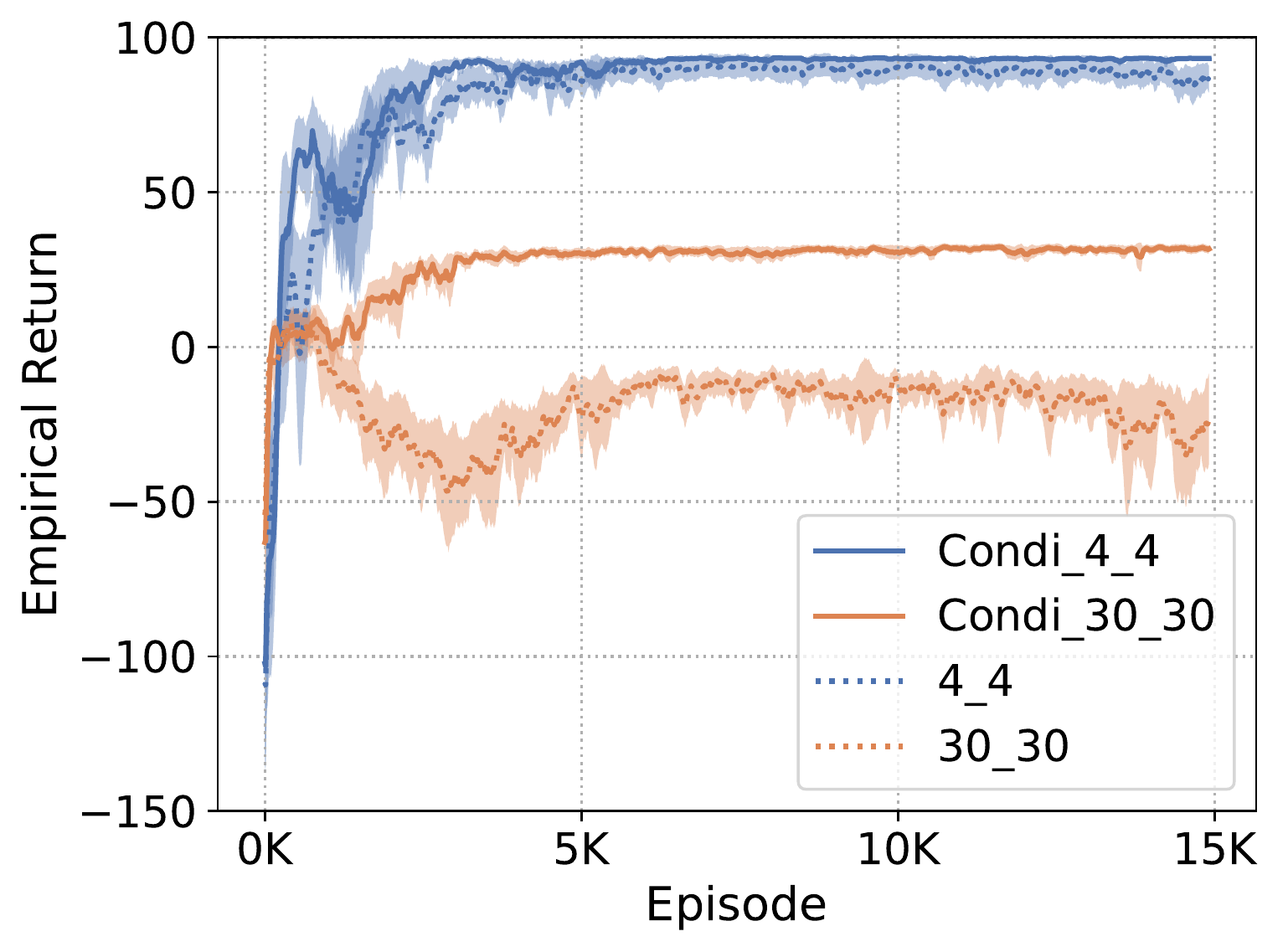}
        \vspace{-2mm}
        \caption{Comparison of learning via conditional (condi) prediction vs unconditional one}
        \vspace{-1mm}
        \label{sc_hc}
    \end{subfigure}
    \qquad
    \begin{subfigure}{.47\textwidth}
        \centering
        \captionsetup{justification=centering}
        \includegraphics[height=3.2cm]{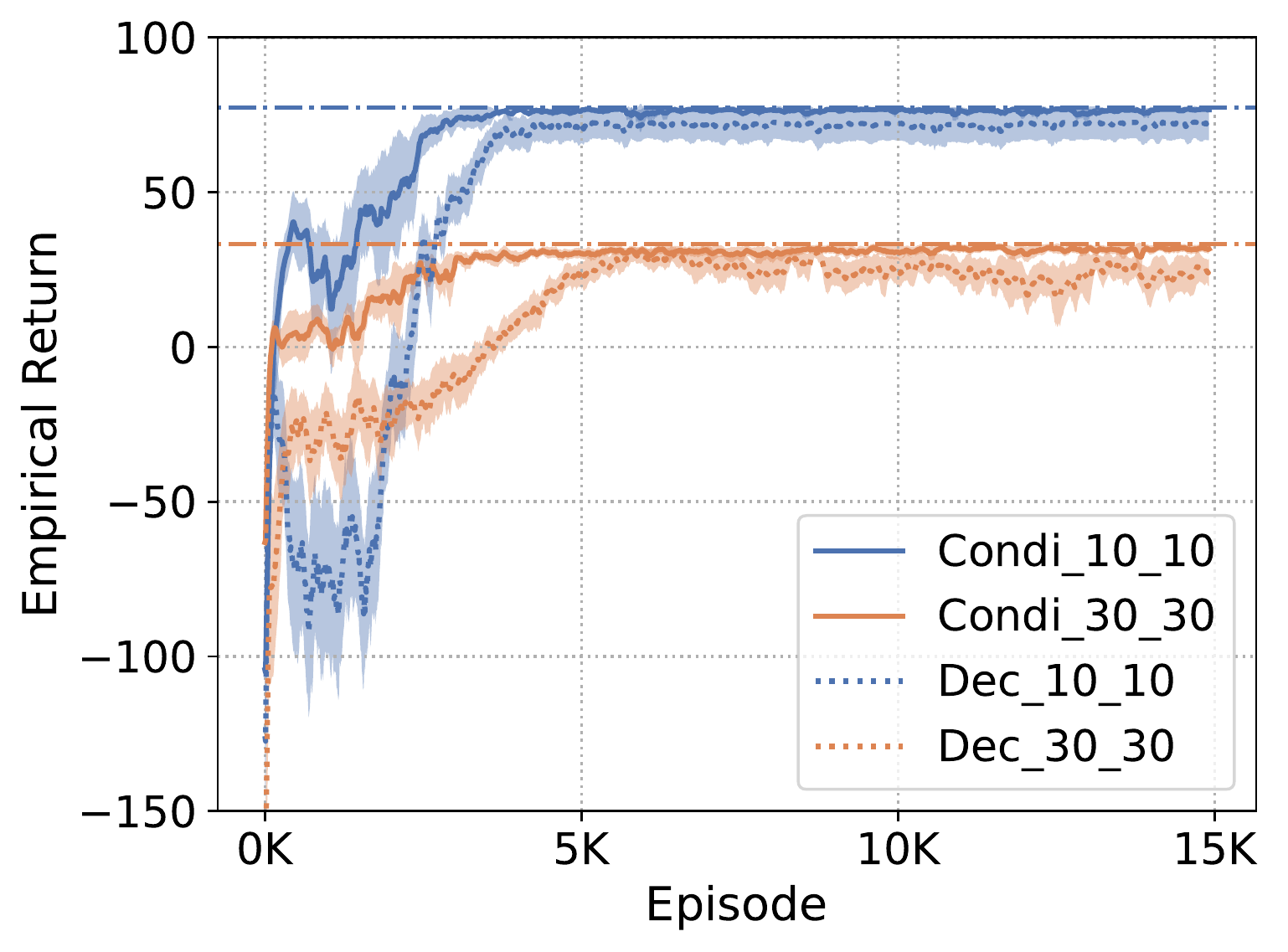}
        \vspace{-2mm}
        \caption{Comparison of centralized learner via conditional prediction vs decentralized learner}
        \vspace{-1mm}
        \label{sc_dec}
    \end{subfigure}
    \caption{Performance of centralized learning in box pushing domains under variant world sizes.}
    \vspace{-5mm}
    \label{cen_result}
\end{figure}



Fig.~\ref{sc_hc} indicates that, in the small grid world ($4 \times 4$), the performance of training centralized policy via \textit{unconditional prediction} (Eq.~\ref{uncondi}) can be as good as the \textit{conditional one} (Eq.~\ref{condi}). This is because the length of the macro-actions (e.g. \textbf{\textit{Push}} and \textbf{\textit{Move\_to\_small\_box}}) is very short, so agents have a high chance to start or end the macro-actions simultaneously. However, in the larger domains, as the asynchronous starting or ending of the macro-actions among the robots becomes more and more dominant, \textit{conditional prediction} is able to provide a more accurate estimation on the target Q-value for training. This is why the conditional method outperforms the unconditional 
one under the grid world size $30 \times 30$. Fig.~\ref{sc_dec} demonstrates that the centralized learner can always learn the best policy and converge to the optimal value (dash-dot line) faster than the decentralized one. 

\subsection{Evaluations in Warehouse Tool Delivery Domain}
\label{WTD}

\begin{wrapfigure}{R}{.4\textwidth}
    \vspace{-5mm}
    \centering
    \includegraphics[height=3.3cm]{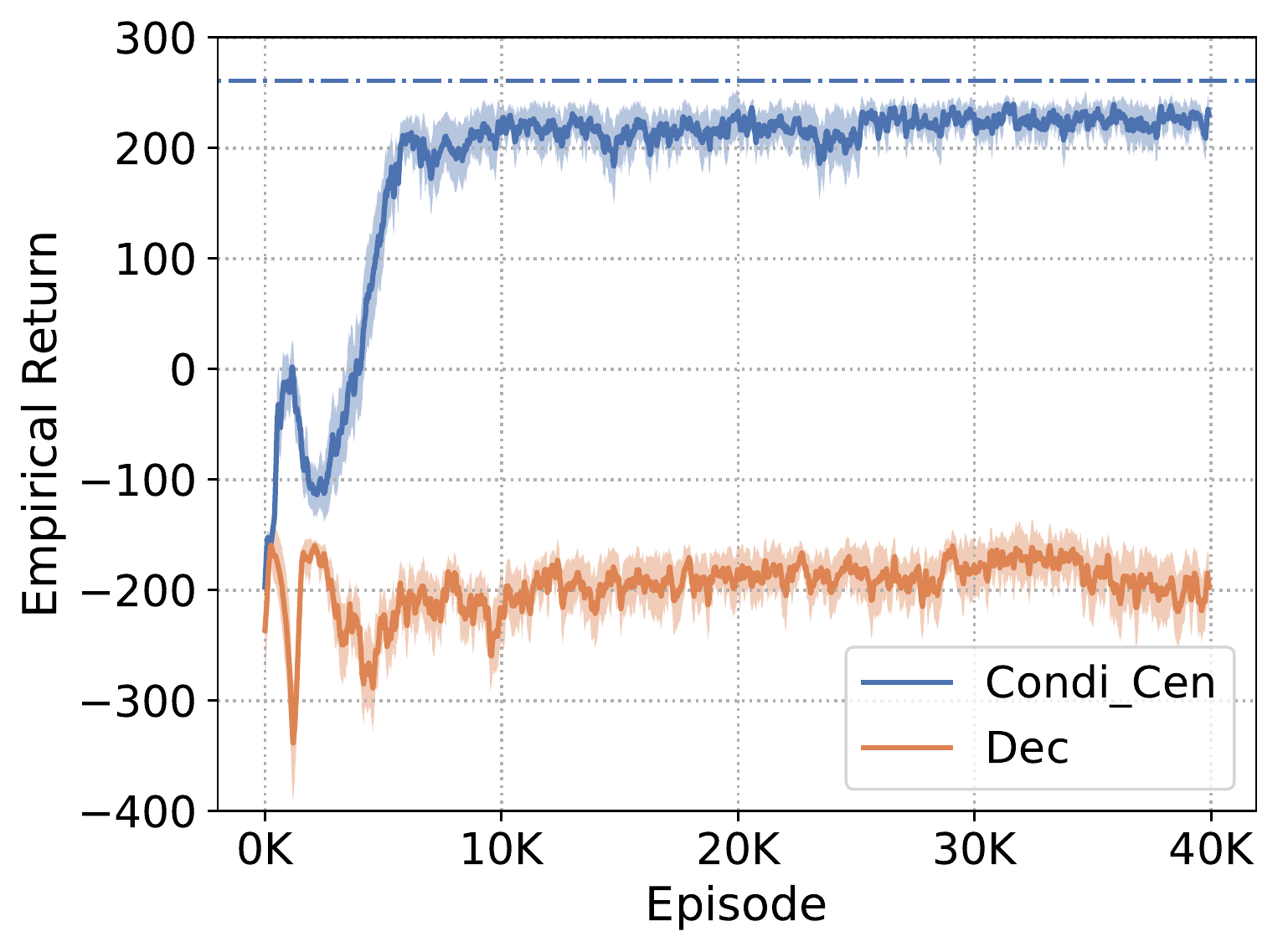}
    \vspace{-2mm}
    \caption{Performance of centralized learning versus decentralized learning under warehouse tool delivery domain.}
    \vspace{-2mm}
    \label{result_wtd}
\end{wrapfigure}

\begin{figure}[h!]
    \centering
    \vspace{-1mm}
    \begin{subfigure}{.15\textwidth}
        \centering
        \vspace{-1mm}
        \includegraphics[height=1.6cm]{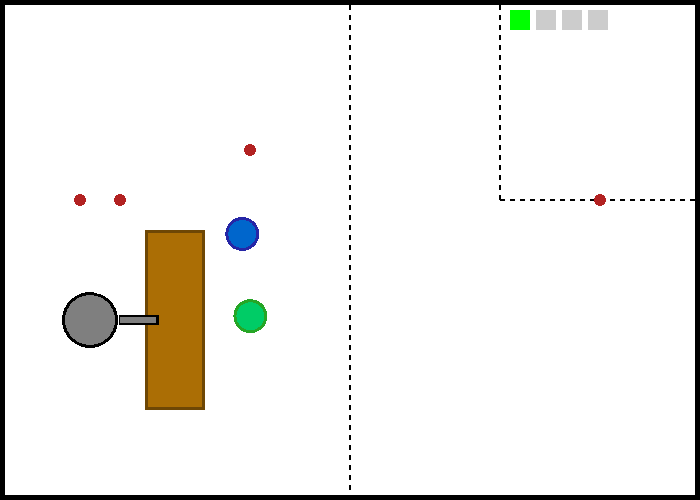}
        \caption{}
        \label{s1_1}
    \end{subfigure}
    ~
    \begin{subfigure}{.15\textwidth}
        \centering
        \vspace{-1mm}
        \includegraphics[height=1.6cm]{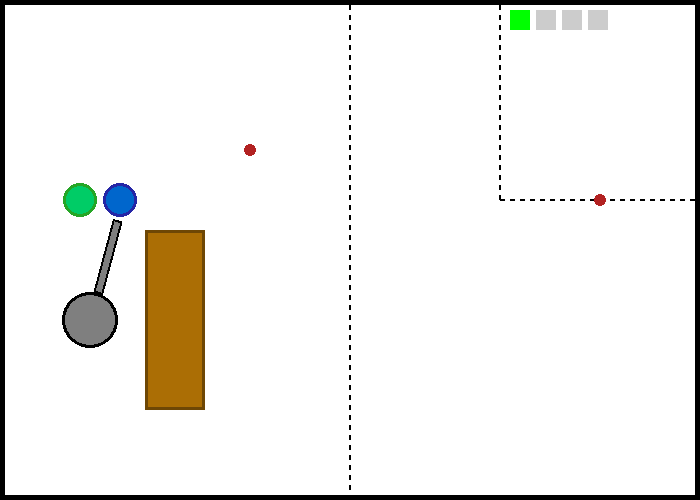}
        \caption{}
        \label{s1_2}
    \end{subfigure}
    ~
    \begin{subfigure}{.15\textwidth}
        \centering
        \vspace{-1mm}
        \includegraphics[height=1.6cm]{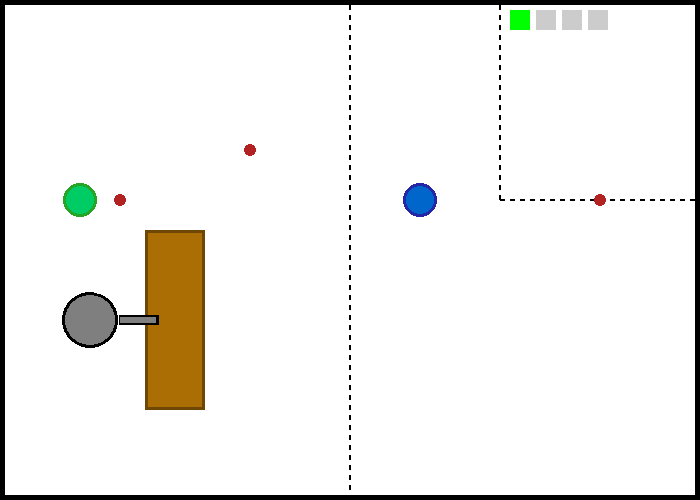}
        \caption{}
        \label{s1_3}
    \end{subfigure}
    ~
    \begin{subfigure}{.15\textwidth}
        \centering
        \vspace{-1mm}
        \includegraphics[height=1.6cm]{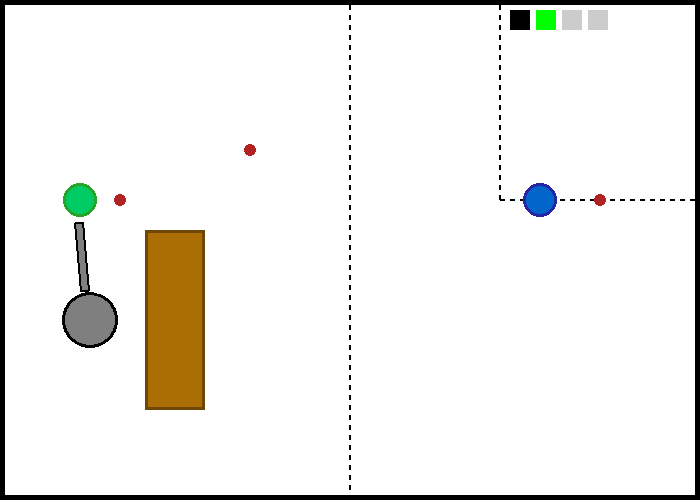}
        \caption{}
        \label{s1_4}
    \end{subfigure}
    ~
    \begin{subfigure}{.15\textwidth}
        \centering
        \vspace{-1mm}
        \includegraphics[height=1.6cm]{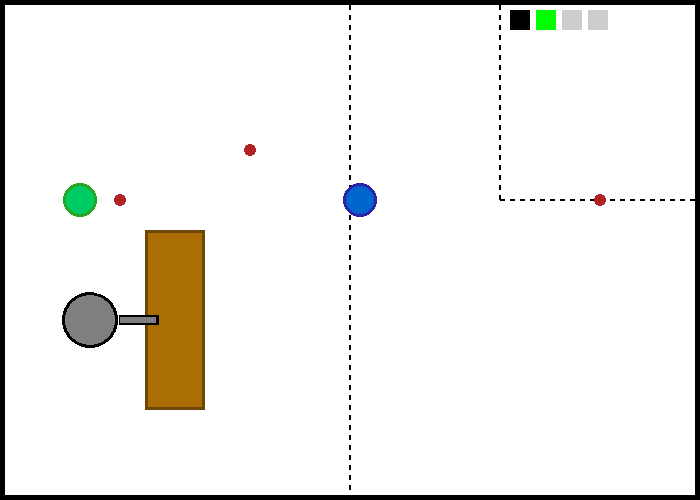}
        \caption{}
        \label{s1_5}
    \end{subfigure}
    ~
    \begin{subfigure}{.15\textwidth}
        \centering
        \vspace{-1mm}
        \includegraphics[height=1.6cm]{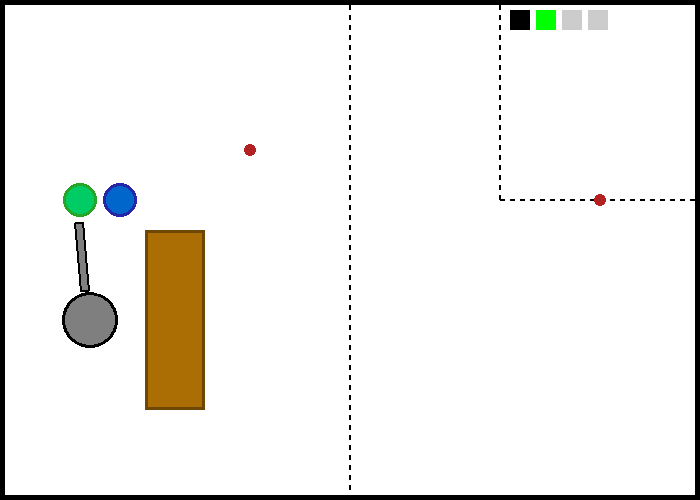}
        \caption{}
        \label{s1_6}
    \end{subfigure}
    \caption{Behaviors of running a centralized policy trained with Turtlebot's velocity $v=0.6$. (a) Fetch robot searches for the first tool for the human while Turtlebots move towards the table; (b-c) One Turtlebot gets the tool from Fetch robot and then delivers it to the human, meanwhile, Fetch starts to search for the second tool; (d) The human obtains the correct tool and moves on the next step, while Fetch passes the second tool to another Turtlebot; (e) The green Turtlebot keeps staying there waiting for the last tool, because delivering two tools together on its own is quicker than letting blue Turtlebot deliver the third one. (f) Green Turtlebot gets the third tool from Fetch, and finishes the entire delivery task in the end (referred to supplementary).}
    \label{WTD1}
    \vspace{-4mm}
\end{figure}
\begin{figure}[h!]
    \centering
    \begin{subfigure}{.15\textwidth}
        \centering
        \includegraphics[height=1.6cm]{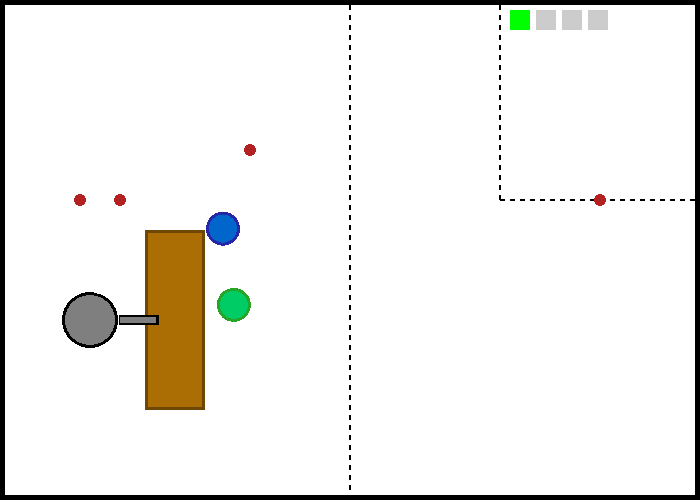}
        \caption{}
        \vspace{-1mm}
        \label{s1}
    \end{subfigure}
    ~
    \begin{subfigure}{.15\textwidth}
        \centering
        \includegraphics[height=1.6cm]{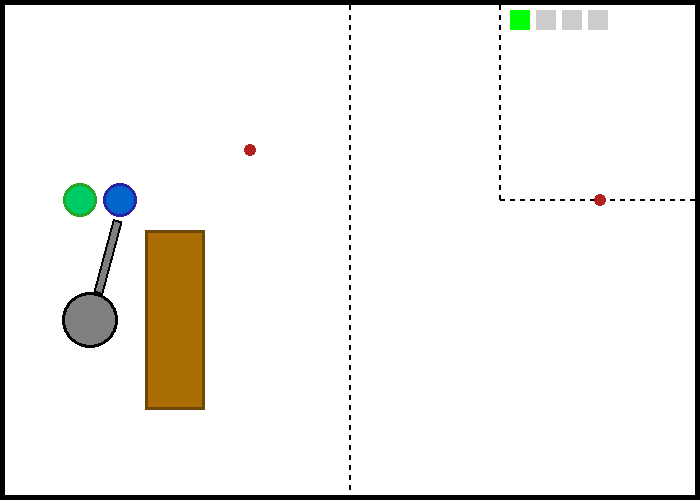}
        \caption{}
        \vspace{-1mm}
        \label{s2}
    \end{subfigure}
    ~
    \begin{subfigure}{.15\textwidth}
        \centering
        \includegraphics[height=1.6cm]{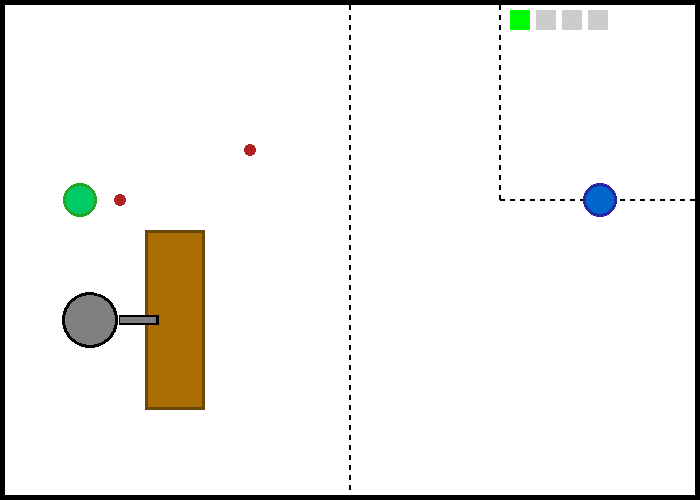}
        \caption{}
        \vspace{-1mm}
        \label{s3}
    \end{subfigure}
    ~
    \begin{subfigure}{.15\textwidth}
        \centering
        \includegraphics[height=1.6cm]{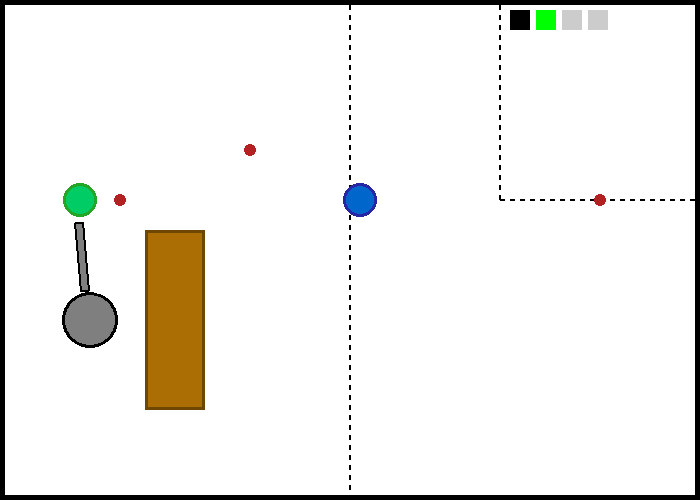}
        \caption{}
        \vspace{-1mm}
        \label{s4}
    \end{subfigure}
    ~
    \begin{subfigure}{.15\textwidth}
        \centering
        \includegraphics[height=1.6cm]{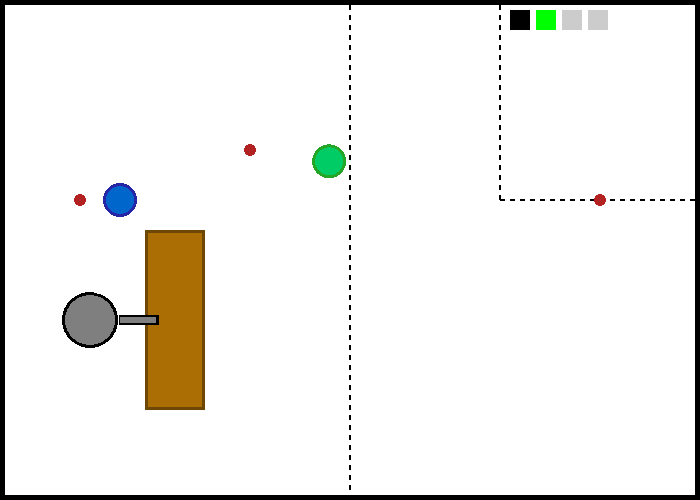}
        \caption{}
        \vspace{-1mm}
        \label{s5}
    \end{subfigure}
    ~
    \begin{subfigure}{.15\textwidth}
        \centering
        \includegraphics[height=1.6cm]{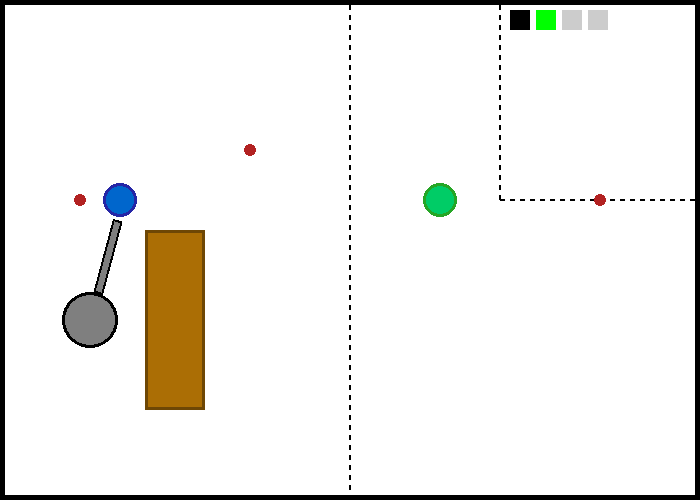}
        \caption{}
        \vspace{-1mm}
        \label{s6}
    \end{subfigure}
    \caption{Behaviors of running the same policy in Fig~\ref{WTD1} with a higher speed $v=0.8$. Differences happen on: (d) after getting the second tool from Fetch robot, (e) the green Turtlebot immediately goes to deliver it and the blue one has already came back to get the third tool. (f) The blue Turtlebot receives the last tool and finally completes the delivery task.  }
    \vspace{-5mm}
    \label{WTD2}
\end{figure}

The optimal collaboration behaviors in this warehouse task depend not only on the time cost of each robot's macro-action execution, but also on how fast the human finishes each step of the task. 
Under the settings introduced in Section~\ref{WTDD}, we performed experiments (using same network architecture as above) on learning both centralized (64 neurons in each MLP layer and 128 neurons in LSTM) and decentralized policies (half of the number of neurons as in the centralized one). 
The result, in Fig.~\ref{result_wtd}, shows that the centralized learner outperforms the decentralized learner, and converges to a value near the optimal one (dash-dot line). 
This is because from Fetch robot's perspective, the reward for delivering a correct tool is very delayed, which depends on the Turtlebots' choices and their moving speeds. Furthermore, a proper delivery requires Fetch to reason about the correct tool even before 
performing cooperating (passing the tool) with the Turtlebots. This is difficult to learn under decentralized training using only local experiences. 
We visualized the trained centralized policy in our simulator to better understand the robots' behaviors, which show that Fetch successfully reasons about the correct tool the human needs per step and cooperates with Turtlebots to finish all deliveries in the optimal way (shown in Fig.~\ref{WTD1}). 
The high robustness of this centralized policy is further demonstrated by being examined under a higher Turtlebot's velocity. The policy generates new collaborative behaviors among robots, which are also optimal with respect to the speed change (shown in Fig.~\ref{WTD2}).  


\section{Conclusion}
\label{sec:conclusion}
\vspace{-5pt}
This paper introduces the first formulation and approach for macro-action-based deep multi-agent reinforcement learning under partial observability. 
Both our decentralized and centralized learners achieve high-quality performance on two benchmark domains. Furthermore, the robots, in the warehouse domain, perform efficient and reasonable cooperation behaviors under the centralized policy. Importantly, the trained policy is naturally robust to the changes on macro-action execution. 
Our formalism and methods open the door for other macro-action-based multi-agent reinforcement learning methods ranging from extensions of other current methods to new approaches and domains. 
As a result, we expect even more scalable learning methods and realistic multi-robot problems. 




\clearpage
\acknowledgments{We thank the reviewers for their time and valuable feedback. We also would like to thank Northeastern University for providing computational resources for generating the results in this paper. This research was funded by ONR
grant N00014-17-1-2072, NSF award 1734497 and 
an Amazon Research Award (ARA). 

}


\bibliography{ref}  

\clearpage
\section*{Supplemental: \thetitle }
\label{sec:supplementary}

\emph{In this section, we present more experiment results under Capture Target, Box Pushing and Warehouse Tool Delivery domains. All the plots are the averaged episodic evaluation returns (evaluation performed every 10 training episodes) over 40 runs with standard error, and smoothed by averaging over 10 neighbors.}

\subsection{Macro-Actions Behaviours in Capture Target Domain}

In this domain, agent's location is fully observable, but the target is flickering with probability $0.3$. Each step, the target (cross) randomly moves along five directions: \textit{up, down, left, right}, or \textit{stay}. Each robot (green or blue circle) has two macro-actions: \textbf{\textit{Move\_to\_Target}}, navigates the robot towards the target and keeps updating the target's location according to the low-level observation; This macro-action will not terminate until reaching the latest observed target's position. Each primitive movement under this macro-action has a transition noisy 0.1.  Note that if the target is flicked, it will continue moving towards the previously observed one; \textbf{\textit{Stay}}, one step macro-action. Only a terminal reward $+1$ can be obtained when the two robots are at the same grid cell with the target simultaneously. Finally, when a robot crosses a border, it is wrapped around and placed on the opposite border in the same row or column. 

\begin{figure}[h!]
    \captionsetup[subfigure]{position=b}
    \centering
    \begin{subfigure}{.45\textwidth}
        \centering
        \includegraphics[height=3.5cm]{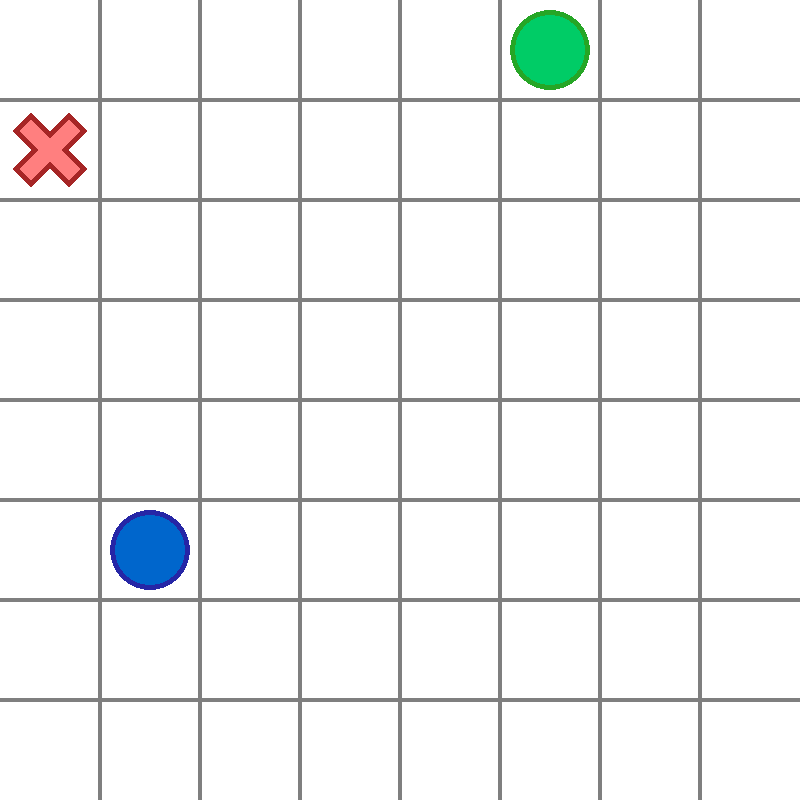}
        \caption{Initial setting with random locations for the target and two robots.}
        \label{ctma_a}
    \end{subfigure}
    \qquad
    \begin{subfigure}{.45\textwidth}
        \vspace{-2mm}
        \centering
        \includegraphics[height=3.5cm]{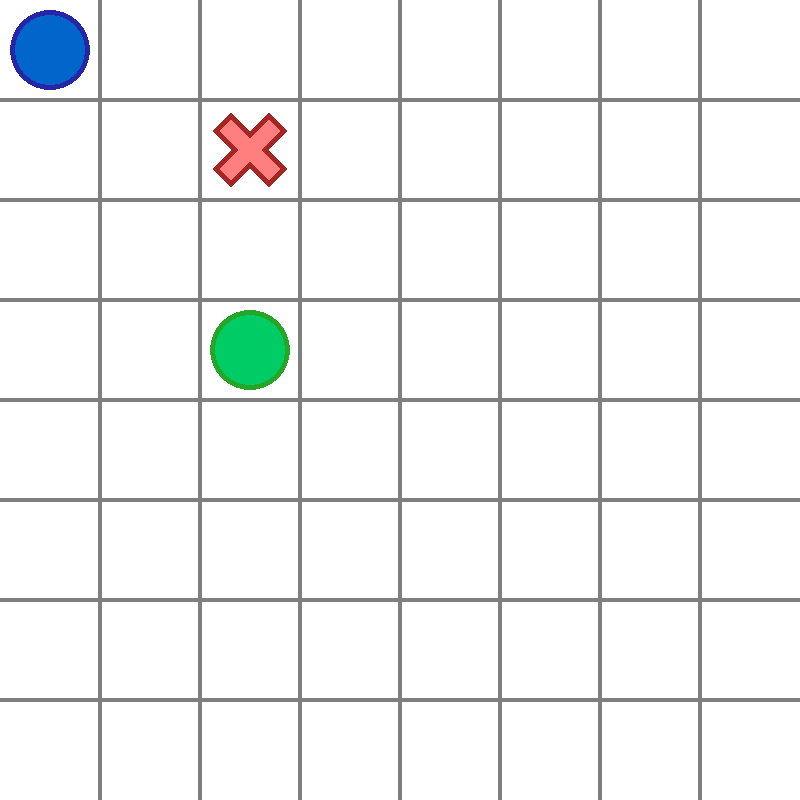}
        \caption{After several executions of \textbf{\textit{Move\_to\_Target}}, two robots get close to the Target.}
        \label{ctma_b}
    \end{subfigure}
    \qquad
    \begin{subfigure}{.45\textwidth}
        \vspace{2mm}
        \centering
        \includegraphics[height=3.5cm]{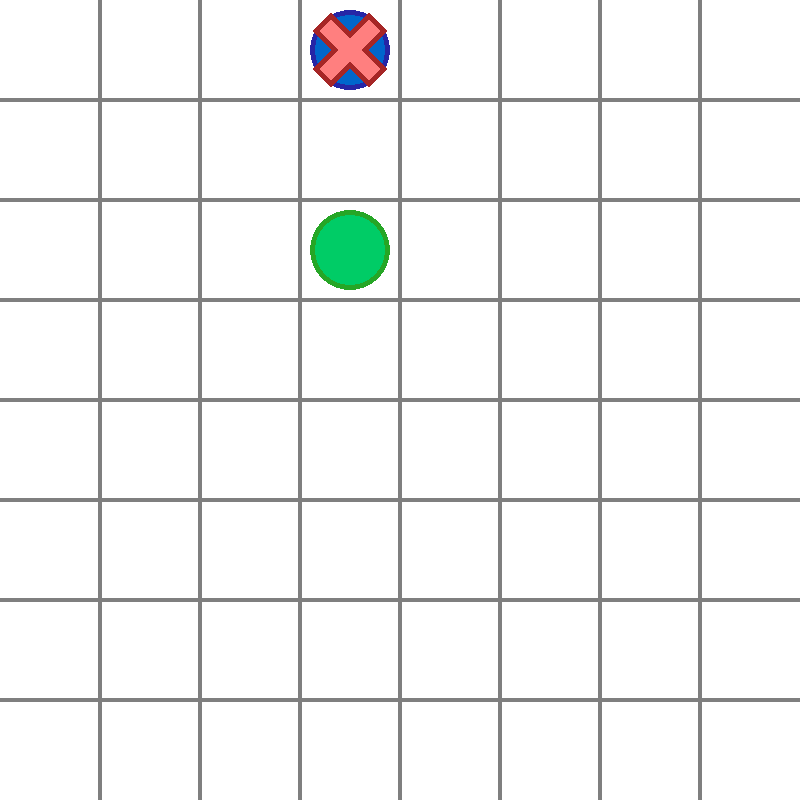}
        \caption{One robot captures the target, but no reward issued and episode keeps running.}
        \label{ctma_c}
    \end{subfigure}
    \qquad
    \begin{subfigure}{.45\textwidth}
        \vspace{2mm}
        \centering
        \includegraphics[height=3.5cm]{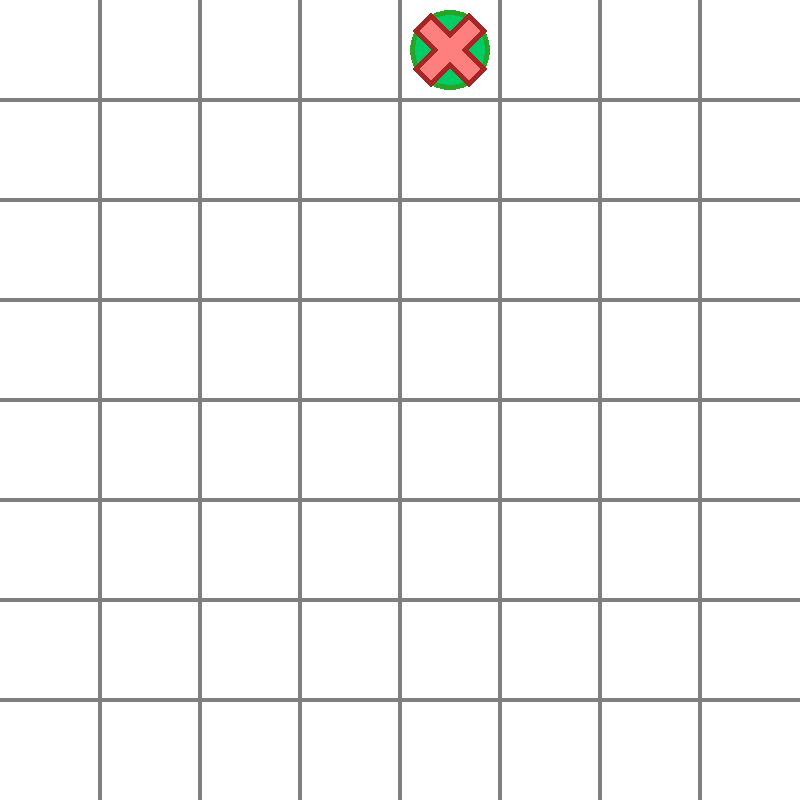}
        \caption{Two robots successfully capture the target together resulting with the terminal $+1$ reward.}
        \label{ctma_d}
    \end{subfigure}
    \qquad
    \caption{Visualization of the behaviours while running decentralized macro-actions-based policy in the Capture Target domain.}
    \vspace{-2mm}
    \label{ctbehave}
\end{figure}

\newpage
\subsection{Macro-Action-Based vs Primitive-action-Based Decentralized Learning in Capture Target Domain}

\begin{figure}[h]
    \vspace{-2mm}
	\centering
    \begin{subfigure}{.29\textwidth}
        \centering
        \includegraphics[height=2.8cm]{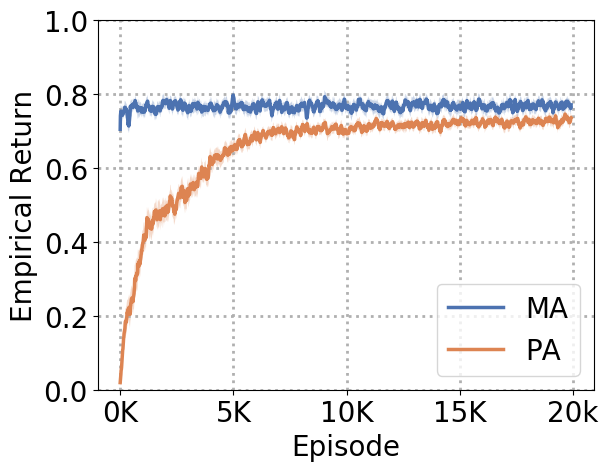}
        \vspace{-1mm}
        \caption{4x4}
        \label{ctma_a}
    \end{subfigure}
    ~
    \begin{subfigure}{.29\textwidth}
        \centering
        \includegraphics[height=2.8cm]{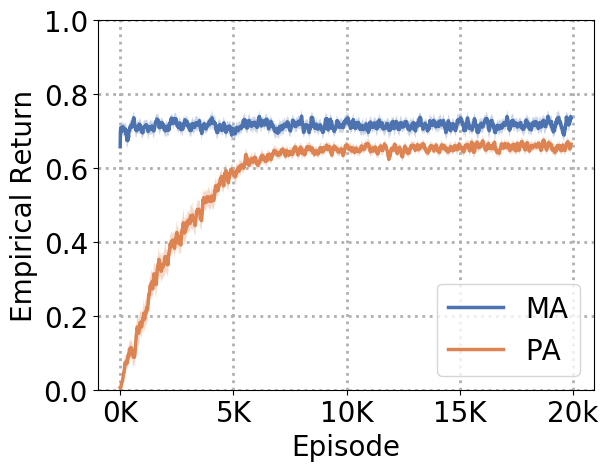}
        \vspace{-1mm}
        \caption{5x5}
        \label{ctma_b}
    \end{subfigure}
    ~
    \begin{subfigure}{.29\textwidth}
        \centering
        \includegraphics[height=2.8cm]{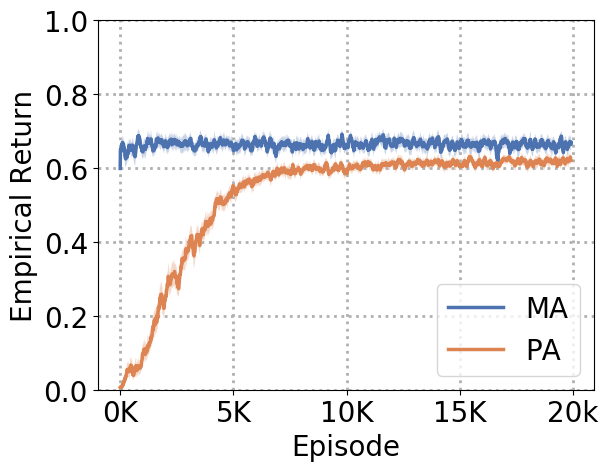}
        \vspace{-1mm}
        \caption{6x6}
        \label{ctma_c}
    \end{subfigure}
    ~
    \begin{subfigure}{.29\textwidth}
        \centering
        \includegraphics[height=2.8cm]{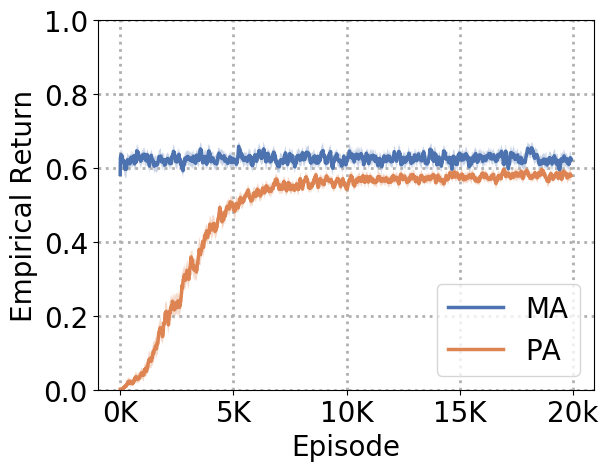}
        \vspace{-1mm}
        \caption{7x7}
        \label{ctma_a}
    \end{subfigure}
    ~
    \begin{subfigure}{.29\textwidth}
        \centering
        \includegraphics[height=2.8cm]{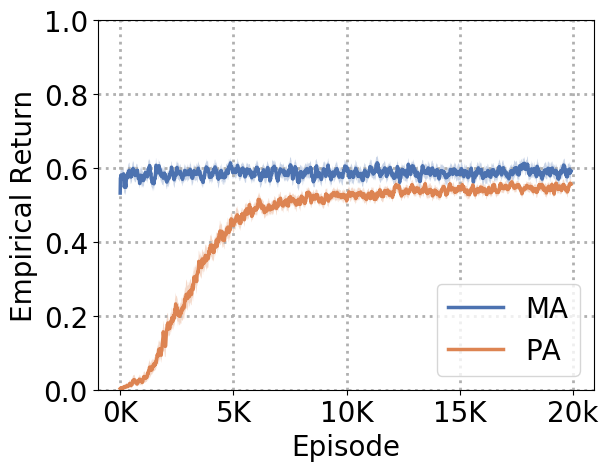}
        \vspace{-1mm}
        \caption{8x8}
        \label{ctma_b}
    \end{subfigure}
    ~
    \begin{subfigure}{.29\textwidth}
        \centering
        \includegraphics[height=2.8cm]{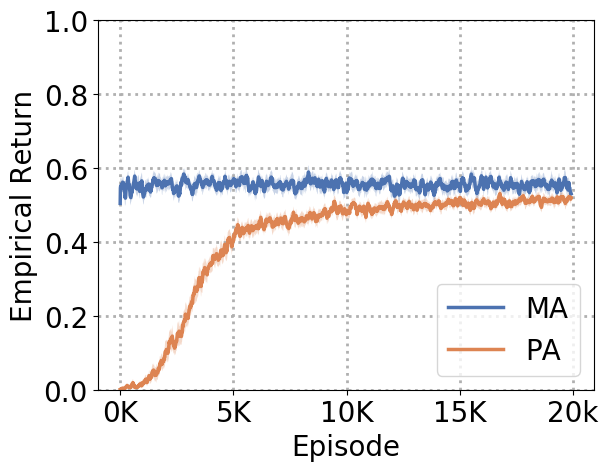}
        \vspace{-1mm}
        \caption{9x9}
        \label{ctma_c}
    \end{subfigure}
    ~
    \begin{subfigure}{.29\textwidth}
        \centering
        \includegraphics[height=2.8cm]{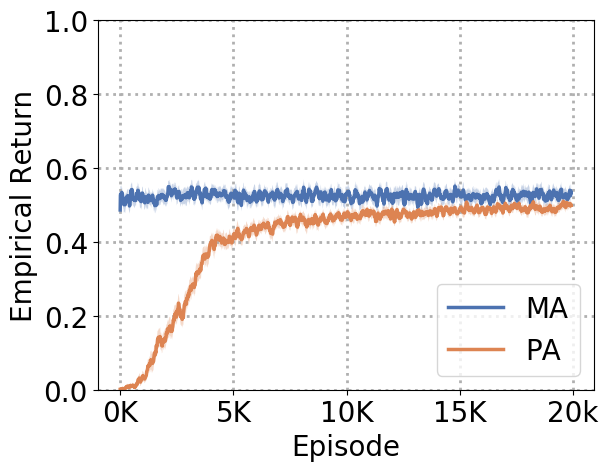}
        \vspace{-1mm}
        \caption{10x10}
        \label{ctma_a}
    \end{subfigure}
    ~
    \begin{subfigure}{.29\textwidth}
        \centering
        \includegraphics[height=2.8cm]{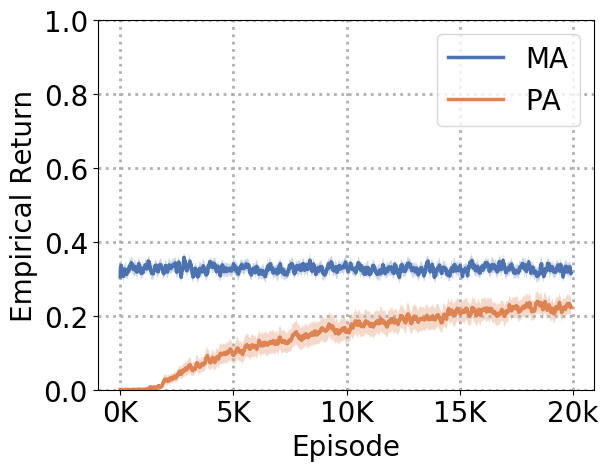}
        \vspace{-1mm}
        \caption{20x20}
        \label{ctma_b}
    \end{subfigure}
    ~
    \begin{subfigure}{.29\textwidth}
        \centering
        \includegraphics[height=3cm]{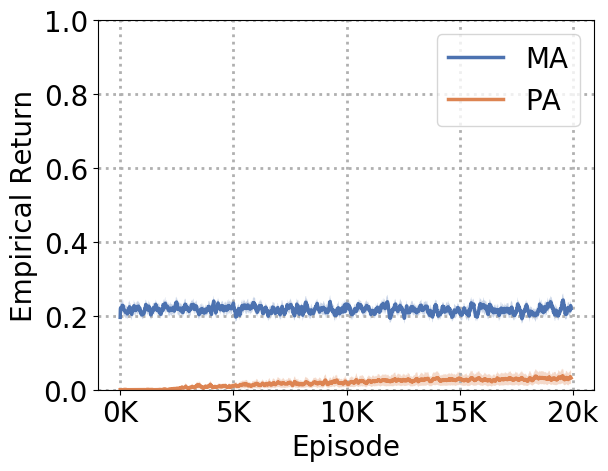}
        \vspace{-1mm}
        \caption{30x30}
        \label{ctma_c}
    \end{subfigure}
    \vspace{-2mm}
    \caption{Comparisons of learning decentralized macro-action (MA) policy and primitive-action (PA) policy in capture target domain under variant gird world spaces.}
    \label{6.2}
\end{figure}

Given the macro-actions, this domain becomes quite simple. 
The results, though, still indicate that learning under macro-actions via our method helps the agents learn better policy and much quicker than learning under primitive-actions.  

\subsection{Optimal Behaviours in Box Pushing Domain with Macro-Actions}
\begin{figure}[h!]
    \centering
    \vspace{-2mm}
    \begin{subfigure}{.3\textwidth}
        \centering
        \captionsetup{justification=centering}
        \includegraphics[height=3cm]{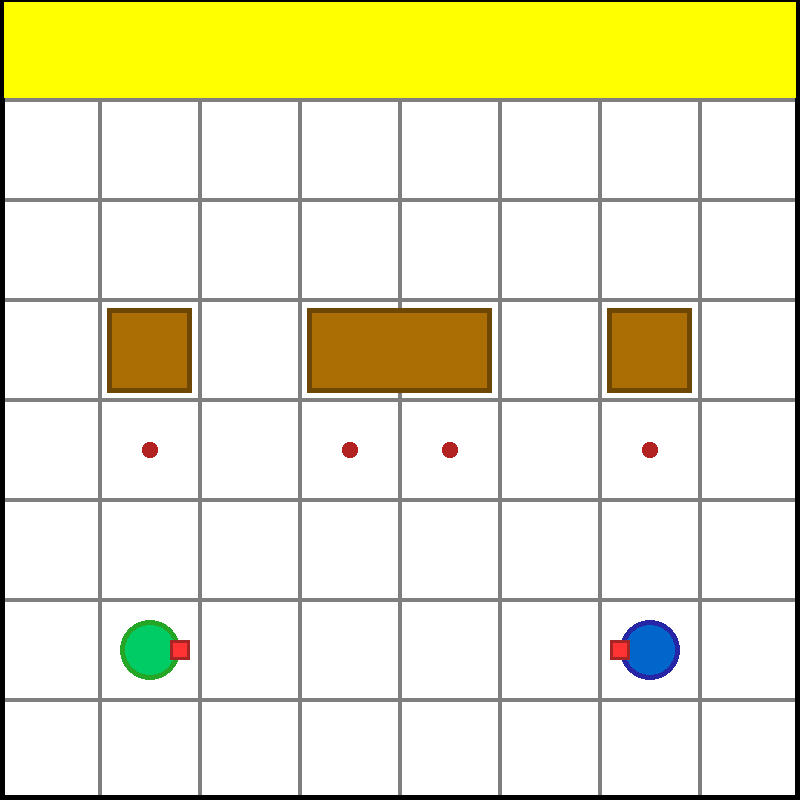}
        \caption{Initial setting of each episode in a $8\times8$ grid world.}
        \label{ctma_a}
    \end{subfigure}
    ~
    \begin{subfigure}{.3\textwidth}
        \centering
        \captionsetup{justification=centering}
        \includegraphics[height=3cm]{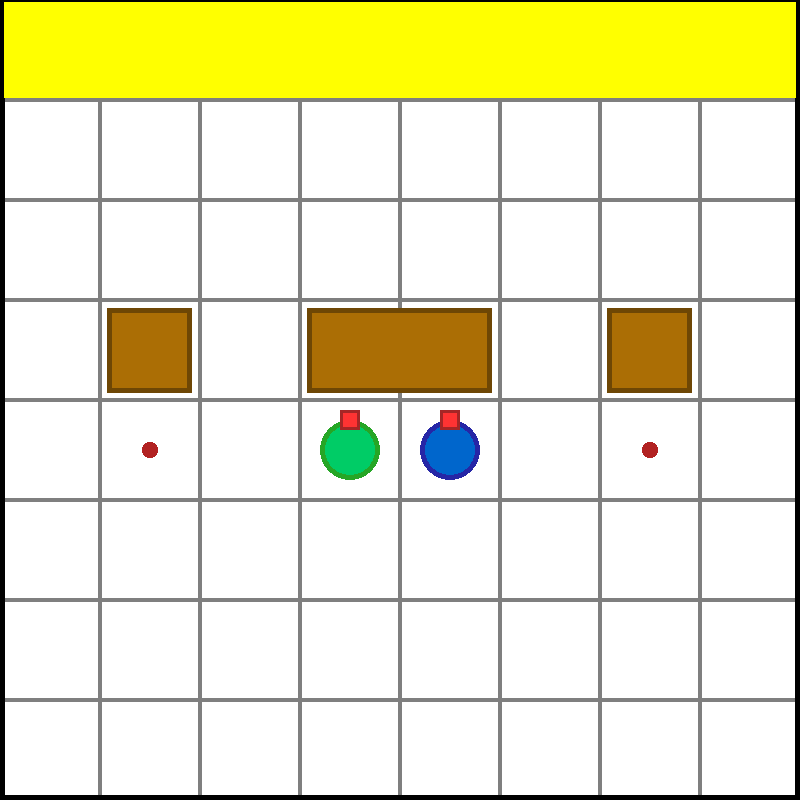}
        \caption{Two robots cooperatively move to the big box.}
        \label{ctma_b}
    \end{subfigure}
    ~
    \begin{subfigure}{.3\textwidth}
        \centering
        \captionsetup{justification=centering}
        \includegraphics[height=3cm]{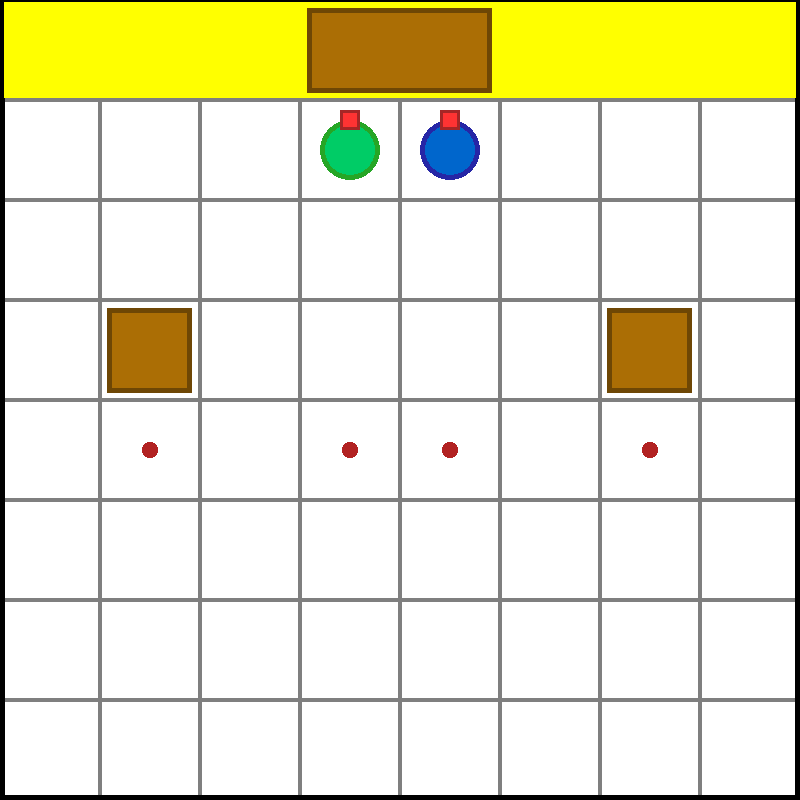}
        \caption{Two robots push the big box to the goal area.}
        \label{ctma_c}
    \end{subfigure}
    \vspace{-1mm}
    \caption{Visualization of the optimal macro-action-based collaboration behaviors learned using our methods in the Box Pushing Domain.}
    \vspace{-2mm}
    \label{bpbehave}
\end{figure}

\subsection{Macro-Action-Based vs Primitive-action-Based Decentralized Learning in Box Pushing Domain}
\begin{figure}[h!]
    \centering
    \begin{subfigure}{.29\textwidth}
        \centering
        \includegraphics[height=3cm]{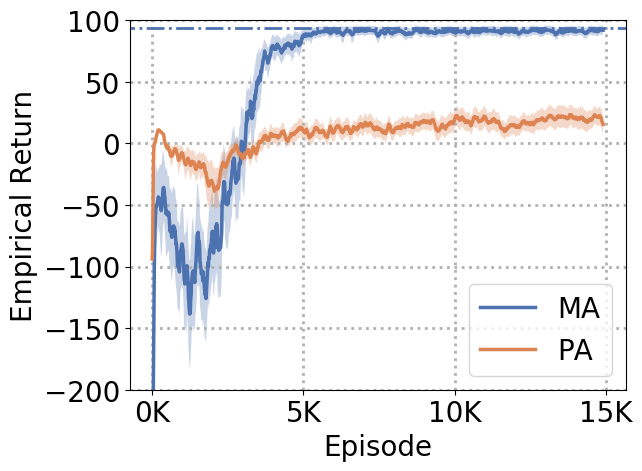}
        \caption{4x4}
        \label{ctma_a}
    \end{subfigure}
    \qquad
    \begin{subfigure}{.29\textwidth}
        \centering
        \includegraphics[height=3cm]{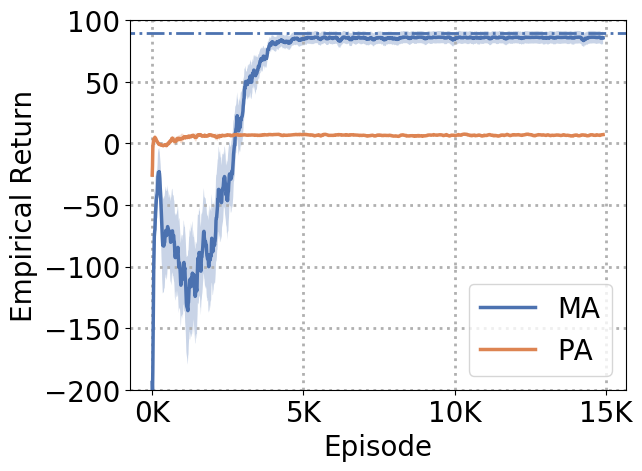}
        \caption{6x6}
        \label{ctma_b}
    \end{subfigure}
    \qquad
    \begin{subfigure}{.29\textwidth}
        \centering
        \includegraphics[height=3cm]{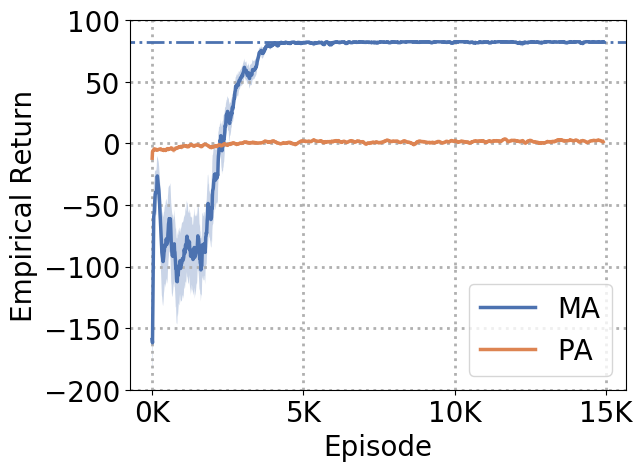}
        \caption{8x8}
        \label{ctma_c}
    \end{subfigure}
    \begin{subfigure}{.29\textwidth}
        \centering
        \includegraphics[height=3cm]{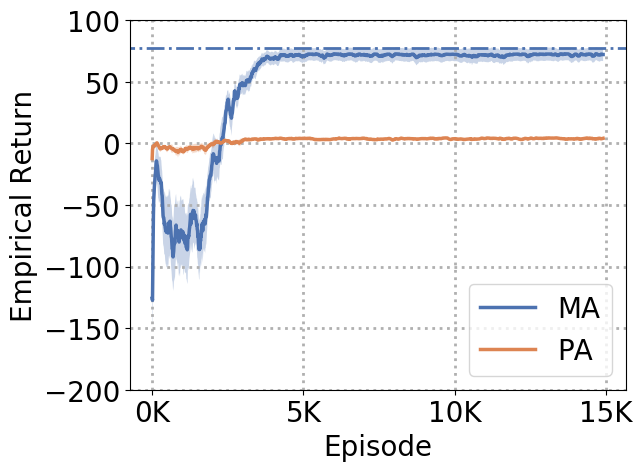}
        \caption{10x10}
        \label{ctma_d}
    \end{subfigure}
    \qquad
    \begin{subfigure}{.29\textwidth}
        \centering
        \includegraphics[height=3cm]{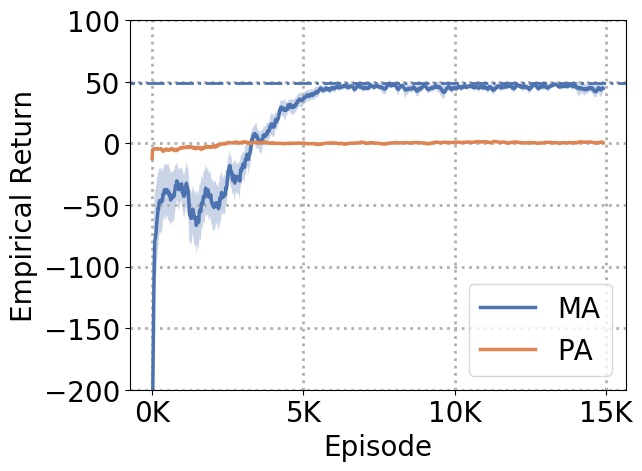}
        \caption{20x20}
        \label{ctma_e}
    \end{subfigure}
    \qquad
    \begin{subfigure}{.29\textwidth}
        \centering
        \includegraphics[height=3cm]{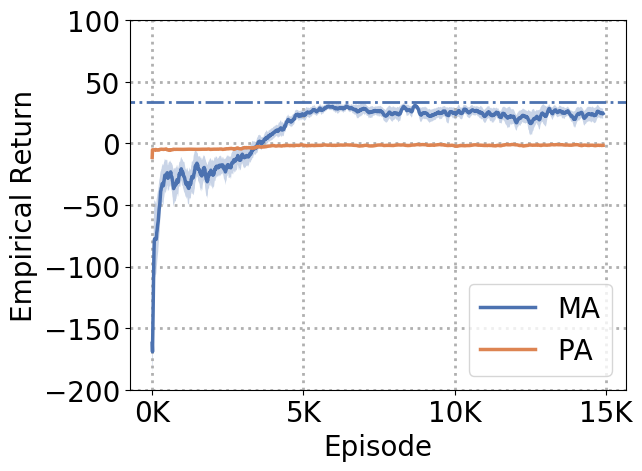}
        \caption{30x30}
        \label{ctma_f}
    \end{subfigure}
    \caption{Comparisons of learning decentralized macro-action (MA) policy versus primitive-action (PA) policy in box pushing domain under variant gird world spaces. 
    The optimal return under each scenario is shown as a dash-dot line.}
    \label{ma_vs_pa_box_pushing}
\end{figure}

The results in Fig.~\ref{ma_vs_pa_box_pushing} show that our approach on learning macro-action-based decentralized policy performs near-optimally, which enables the robots to cooperatively push the big box and does not suffer from the world space increasing. The primitive learner either converges to a local optimum or cannot learn anything in a larger gird world. 

\newpage
\subsection{Macro-Action-Based Centralized Learning in Box Pushing Domain}

The centralized training results shown below demonstrate the advantage of \emph{conditional target prediction} over the \emph{unconditional} one. In the small environment ($4\times4$), training using \emph{unconditional prediction} achieves similar performance to the \emph{conditional one}, but it becomes worse and worse with world space increases. Because, in the larger world space, there are more asynchronous executions among agents, thus there is less accurate estimation provided by the \emph{unconditional one}. It is also interesting to note that, under the middle size world (e.g. $8\times8$), random exploration behavior ($\epsilon-$greedy) reduces the negative influence of \textit{unconditional prediction}. However, the estimation error keeps getting accumulated and finally leads the learning to be worse results.

\begin{figure}[h!]
	\centering
    \begin{subfigure}{.29\textwidth}
        \centering
        \includegraphics[height=3cm]{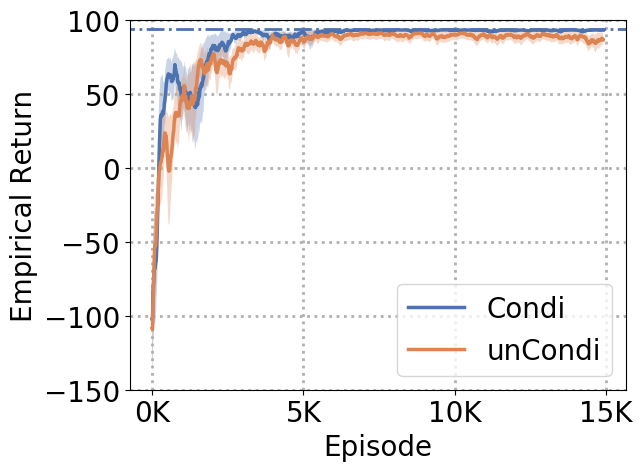}
        \caption{4x4}
        \label{ctma_a}
    \end{subfigure}
    \qquad
    \begin{subfigure}{.29\textwidth}
        \centering
        \includegraphics[height=3cm]{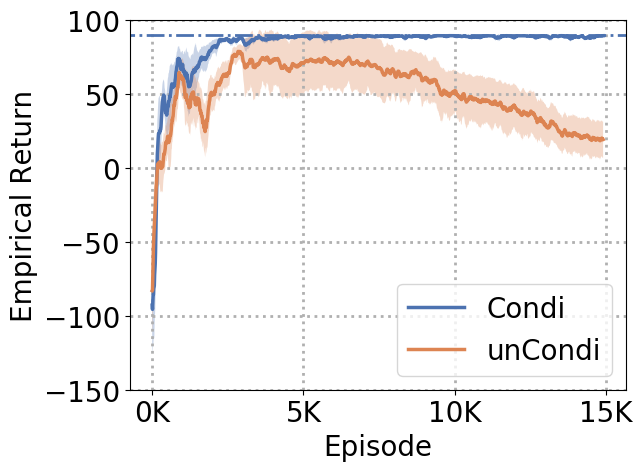}
        \caption{6x6}
        \label{ctma_b}
    \end{subfigure}
    \qquad
    \begin{subfigure}{.29\textwidth}
        \centering
        \includegraphics[height=3cm]{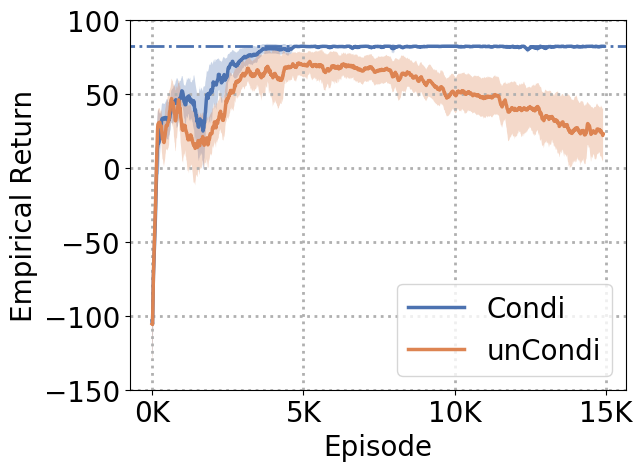}
        \caption{8x8}
        \label{ctma_c}
    \end{subfigure}
    \qquad
    \begin{subfigure}{.29\textwidth}
        \centering
        \includegraphics[height=3cm]{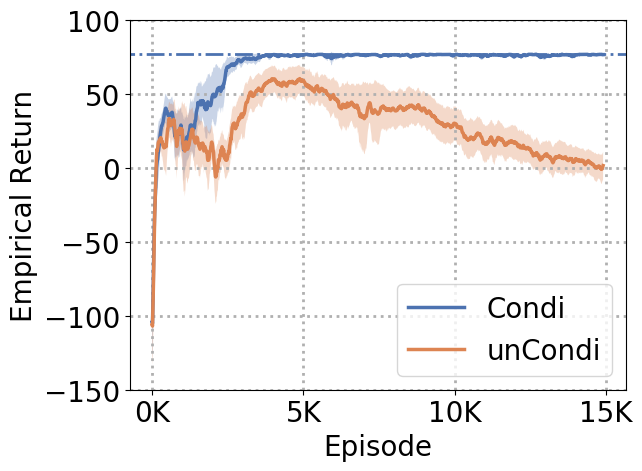}
        \caption{10x10}
        \label{ctma_a}
    \end{subfigure}
    \qquad
    \begin{subfigure}{.29\textwidth}
        \centering
        \includegraphics[height=3cm]{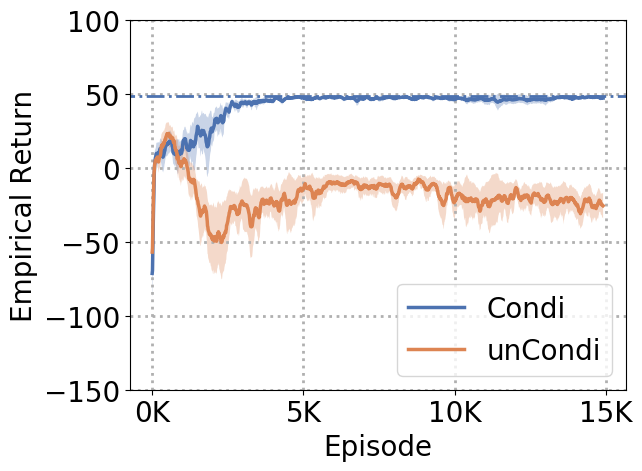}
        \caption{20x20}
        \label{ctma_b}
    \end{subfigure}
    \qquad
    \begin{subfigure}{.29\textwidth}
        \centering
        \includegraphics[height=3cm]{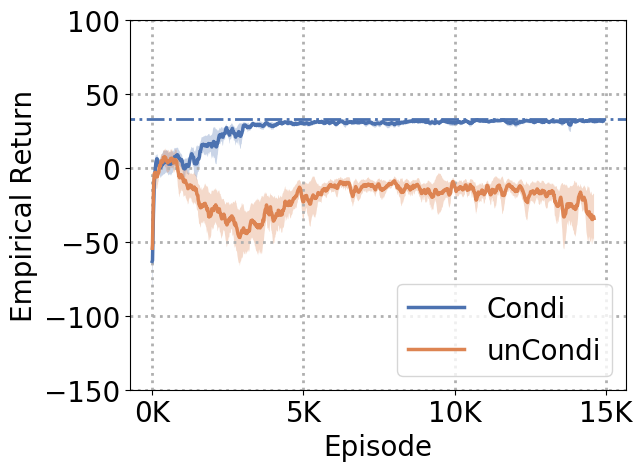}
        \caption{30x30}
        \label{ctma_c}
    \end{subfigure}
    \qquad
    \caption{Comparisons of learning macro-action-based centralized policy via \emph{conditional target prediction} (Condi) versus \emph{unconditional target prediction} (unCondi) in the Box Pushing domain under variant grid world sizes. The dash-dot line represents the corresponding optimal return value.}
    \label{bpcondi_vs_uncondi}
\end{figure}

\subsection{Learning Macro-Action-Based Centralized Policy vs Decentralized Policy in Box Pushing Domain}

\begin{figure}[h!]
	\centering
    \begin{subfigure}{.29\textwidth}
        \centering
        \includegraphics[height=3cm]{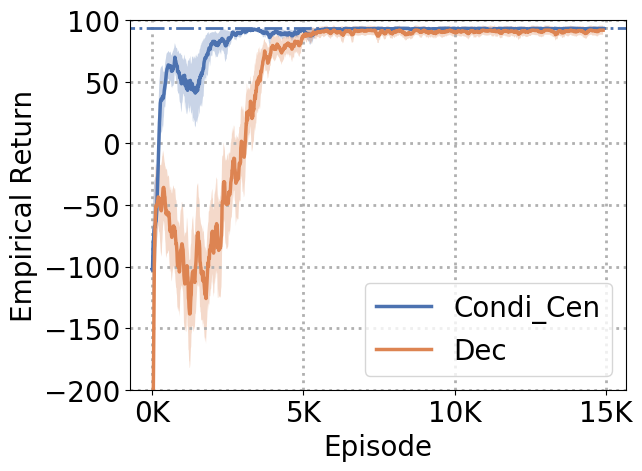}
        \caption{4x4}
        \label{ctma_a}
    \end{subfigure}
    \qquad
    \begin{subfigure}{.29\textwidth}
        \centering
        \includegraphics[height=3cm]{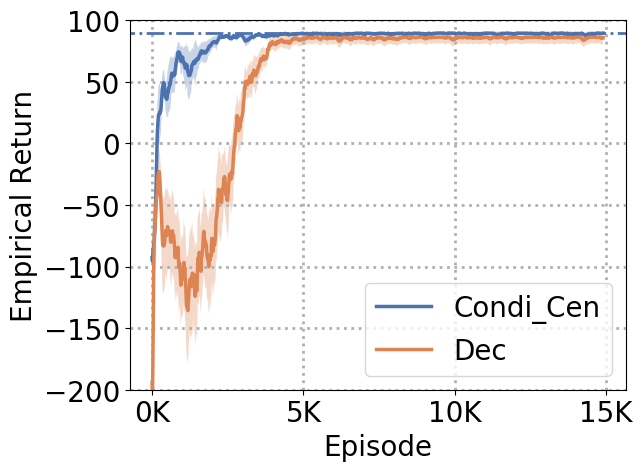}
        \caption{6x6}
        \label{ctma_b}
    \end{subfigure}
    \qquad
    \begin{subfigure}{.29\textwidth}
        \centering
        \includegraphics[height=3cm]{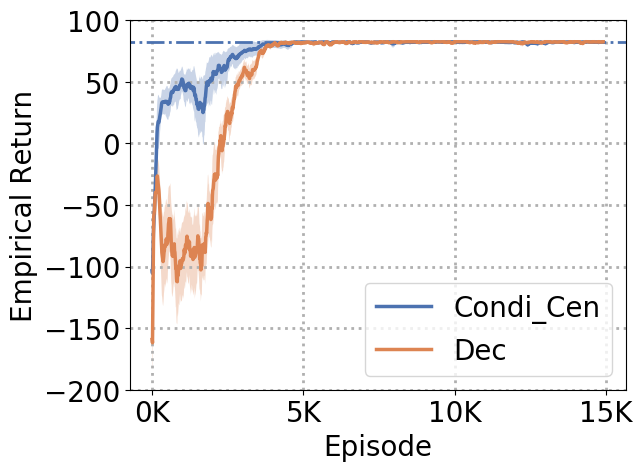}
        \caption{8x8}
        \label{ctma_c}
    \end{subfigure}
    \qquad
    \begin{subfigure}{.29\textwidth}
        \centering
        \includegraphics[height=3cm]{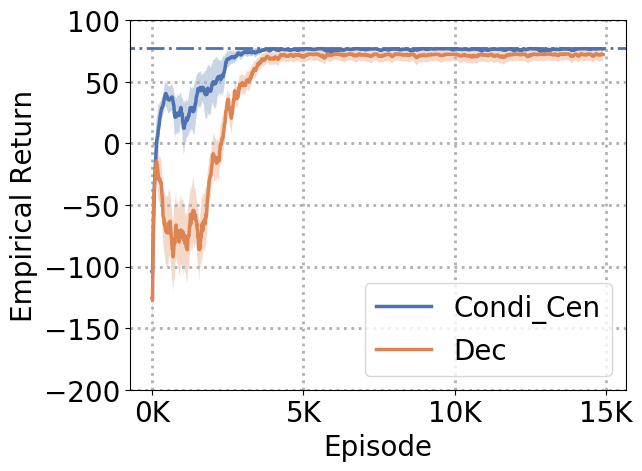}
        \caption{10x10}
        \label{ctma_a}
    \end{subfigure}
    \qquad
    \begin{subfigure}{.29\textwidth}
        \centering
        \includegraphics[height=3cm]{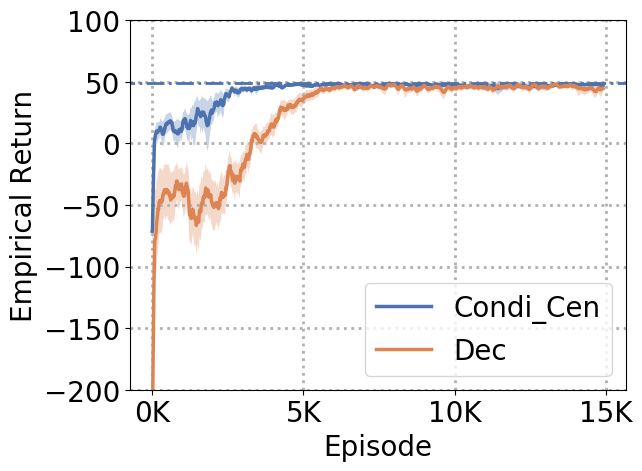}
        \caption{20x20}
        \label{ctma_b}
    \end{subfigure}
    \qquad
    \begin{subfigure}{.29\textwidth}
        \centering
        \includegraphics[height=3cm]{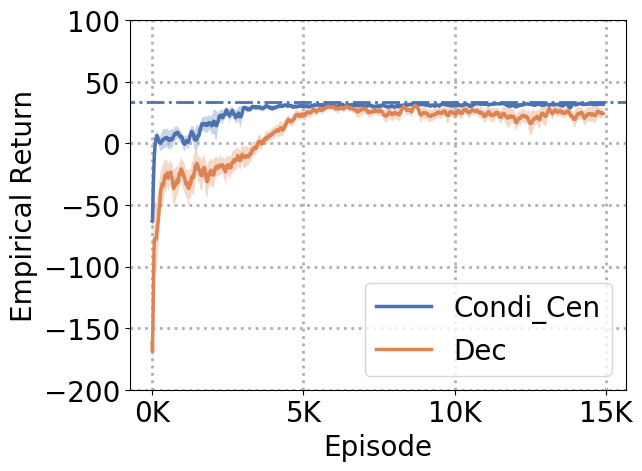}
        \caption{30x30}
        \label{ctma_c}
    \end{subfigure}
    \caption{Comparisons of macro-action-based centralized training via \emph{conditional target prediction} versus decentralized training. Optimal return under each scenario is shown as a dash-dot line.}
    \label{6.6}
\end{figure}

Centralized training receives all the robots' observations as input, which facilitates the robots to learn the optimal collaboration behavior faster than decentralized training that only uses local information. Under all the scenarios in Fig.~\ref{6.6}, the centralized learner can always converge to the optimal value, which further demonstrates the correctness of our approach on learning joint macro-action-value function. 

\newpage
\subsection{Examination of the Trained Centralized Policy in Warehouse Tool Delivery Domain}

\begin{figure}[h!]
    \centering
    \subcaptionbox{Two Turtlebots execute \textbf{\textit{Get\_Tool}} to move towards the table, and Fetch runs \textbf{\textit{Search\_Tool}}$(0)$ to search the first tool for the human.}
        [0.3\linewidth]{\includegraphics[height=2.5cm]{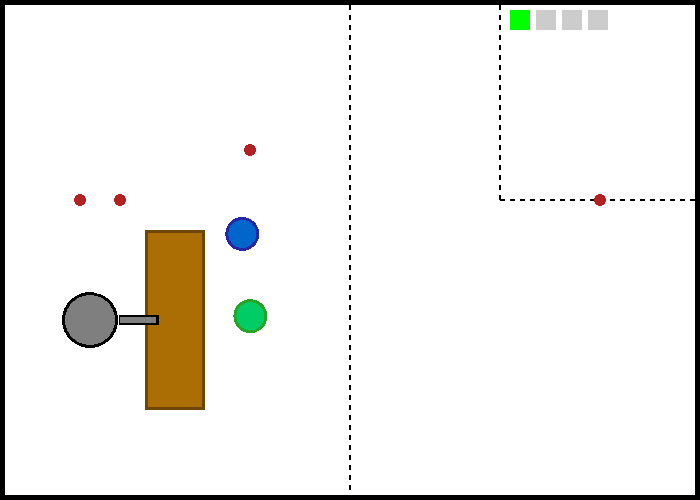}}
    ~
    \subcaptionbox{Two Turtlebots are still under \textbf{\textit{Get\_Tool}} waiting beside the table, and Fetch keeps looking for the first tool.}
        [0.3\linewidth]{\includegraphics[height=2.5cm]{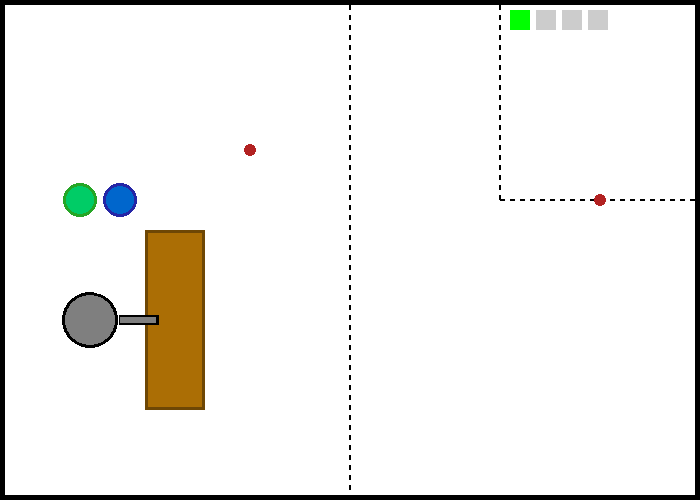}}
    ~
    \subcaptionbox{Fetch executes \textbf{\textit{Pass\_to\_T(1)}} to pass the first tool to Turtelbot\_$1$.}
        [0.3\linewidth]{\includegraphics[height=2.5cm]{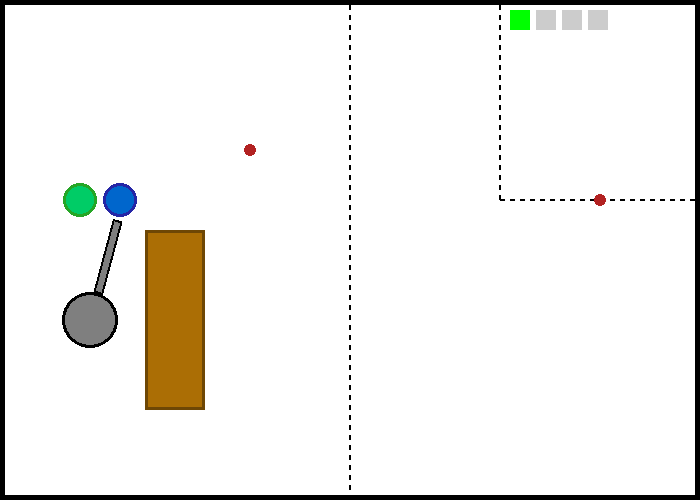}}
    \par\bigskip
    \subcaptionbox{Turtelbot\_$1$ executes \textbf{\textit{Go\_to\_WS}} to deliver the tool to the human, meanwhile, Fetch starts to look for the second tool by running \textbf{\textit{Search\_Tool}}$(1)$}
        [0.3\linewidth]{\includegraphics[height=2.5cm]{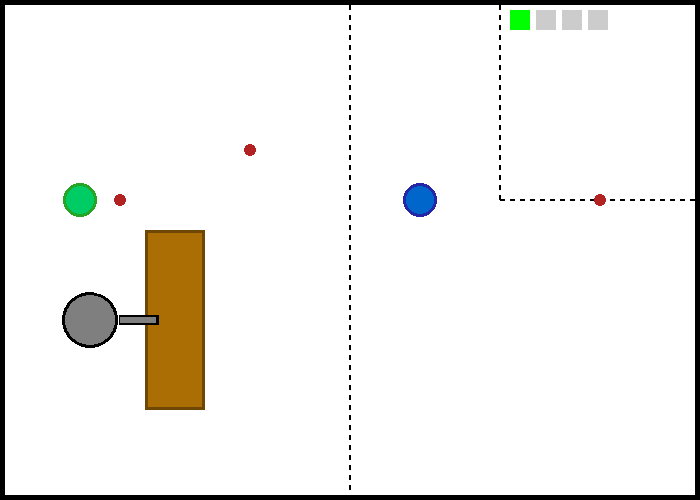}}
    ~
    \subcaptionbox{Turtlebot\_$1$ arrives at the workshop right on time when the human finished the first task's step.}
        [0.3\linewidth]{\includegraphics[height=2.5cm]{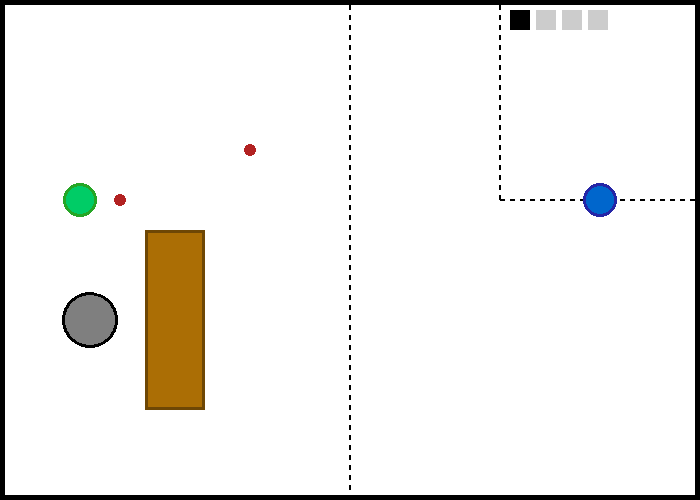}}
    ~
    \subcaptionbox{Human gets tool from Turtelbot\_1 and continues working; Turtlebot\_1 executes \textbf{\textit{Get\_Tool}} to go back tool room; Fetch now runs \textbf{\textit{Pass\_to\_T(1)}} to pass the second tool to the Turtlebot\_$0$.}
        [0.3\linewidth]{\includegraphics[height=2.5cm]{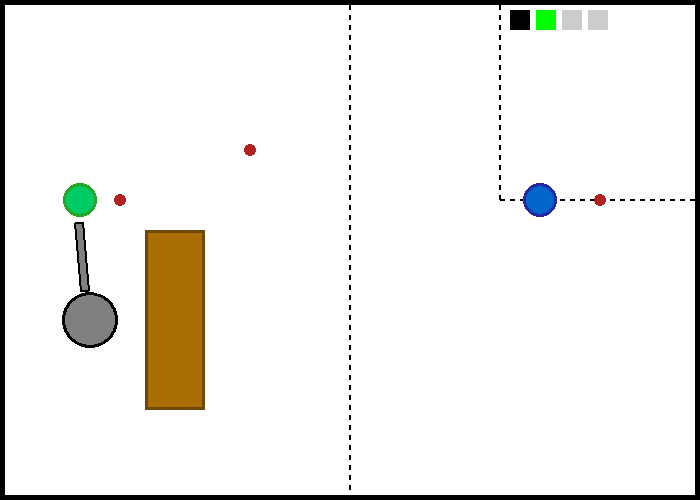}}
    \par\bigskip
    \subcaptionbox{Turtlebot\_$0$ keeps waiting there (under running \textbf{\textit{Get\_Tool}}) for the third tool which Fetch is searching for by running \textbf{\textit{Search\_Tool}}$(2)$.}
        [0.3\linewidth]{\includegraphics[height=2.5cm]{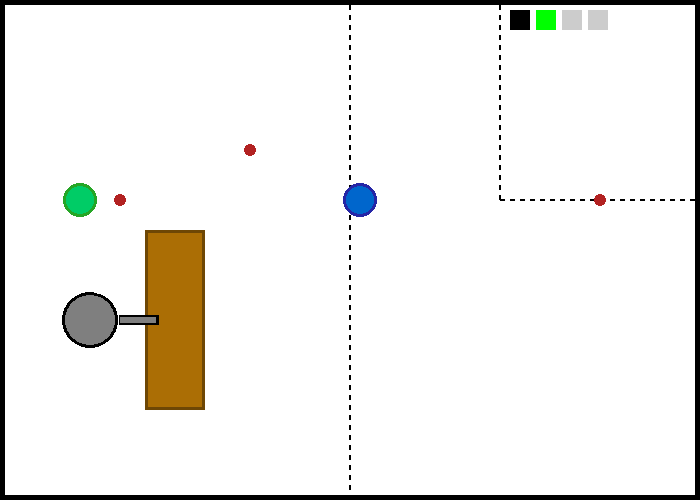}}
    ~
    \subcaptionbox{Fetch executes \textbf{\textit{Pass\_to\_T}}$(0)$ to give the last tool to Turtlebot\_0.}
        [0.3\linewidth]{\includegraphics[height=2.5cm]{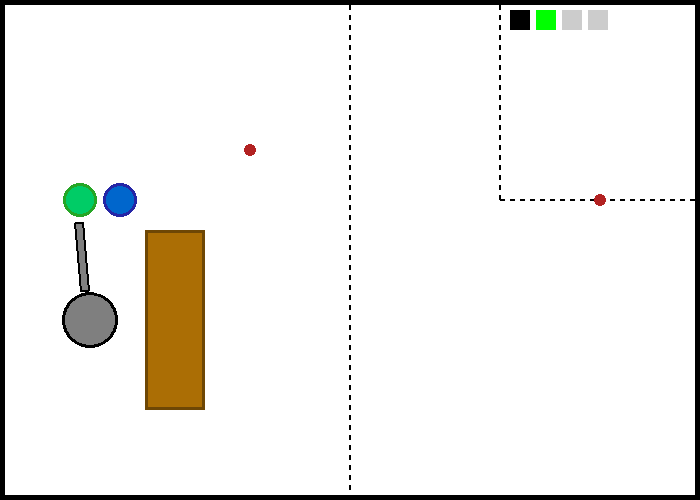}}
    ~
    \subcaptionbox{Turtelbot\_$0$ arrives the workshop when human finishes the second step. Human first gets the tool for the third step from the Turtlebot, and then also obtains the last tool for step 4. Robots finish the entire delivery task.}
        [0.3\linewidth]{\includegraphics[height=2.5cm]{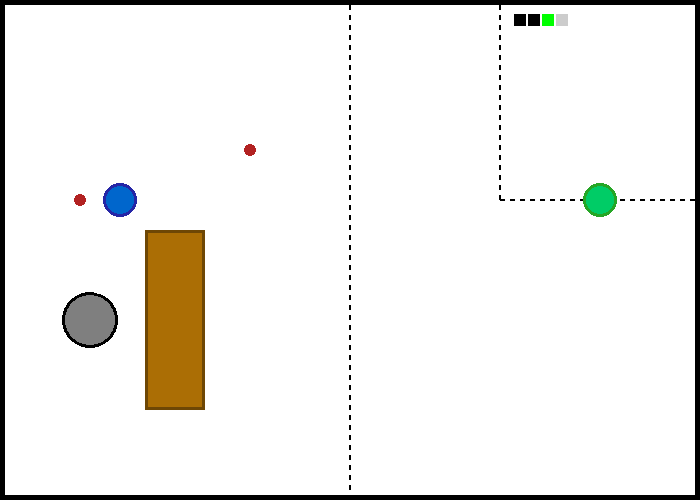}}

    \caption{Visualization of running a centralized policy trained under Turtelbot moving speed $v=0.6$.}
    \label{wtds1}
\end{figure}

In this section, we first show the collaborative behaviors performed by running the trained centralized policy under the Turtlebot moving speed $v=0.6$ (Fig.~\ref{wtds1}). 

\begin{figure}[h!]
    \centering
    \subcaptionbox{Two Turtlebots execute \textbf{\textit{Get\_Tool}} to move towards the table, and Fetch runs \textbf{\textit{Search\_Tool}}$(0)$ to search the first tool for the human.}
        [0.3\linewidth]{\includegraphics[height=2.5cm]{s2_1.png}}
    ~
    \subcaptionbox{Two Turtlebots are still under \textbf{\textit{Get\_Tool}} waiting beside the table, and Fetch keeps looking for the first tool.}
        [0.3\linewidth]{\includegraphics[height=2.5cm]{ss1_2.png}}
    ~
    \subcaptionbox{Fetch executes \textbf{\textit{Pass\_to\_T(1)}} to pass the first tool to Turtelbot\_$1$.}
        [0.3\linewidth]{\includegraphics[height=2.5cm]{s2_2.png}}
    \par\bigskip
    \subcaptionbox{Turtelbot\_$1$ executes \textbf{\textit{Go\_to\_WS}} to deliver the tool to the human, meanwhile, Fetch starts to look for the second tool by running \textbf{\textit{Search\_Tool}}$(1)$. Note that this first delivery is faster than the one shown in Fig.~\ref{wtds1}e.}
        [0.3\linewidth]{\includegraphics[height=2.5cm]{s2_3.png}}
    ~
    \subcaptionbox{Human starts working on step 2 after getting the tool from Turtlebot\_1, and Turtlebot\_1 goes back to tool room. In the meantime, Fetch executes \textbf{\textit{Pass\_to\_T(0)}} to pass the second tool to Turtelbot\_$0$}
        [0.3\linewidth]{\includegraphics[height=2.5cm]{s2_4.png}}
    ~
    \subcaptionbox{Turtlebot\_$0$ immediately leaves for delivering the second tool running \textbf{\textit{Go\_to\_WS}} rather than waiting over there, and Fetch starts looking for the last tool by executing \textbf{\textit{Search\_Tool}}$(2)$.} 
        [0.3\linewidth]{\includegraphics[height=2.5cm]{s2_5.png}}
    \par\bigskip
    \subcaptionbox{Turtlebot\_$0$ delivers the tool human needs for the third step, and Fetch just found the last tool.} 
        [0.3\linewidth]{\includegraphics[height=2.5cm]{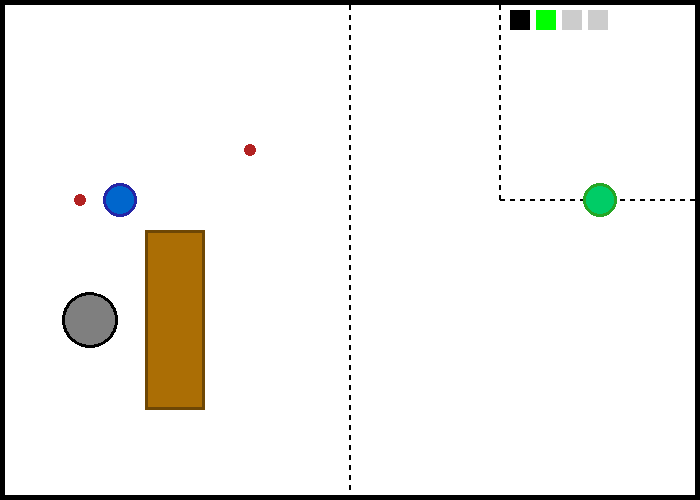}}
    ~
    \subcaptionbox{Fetch executes \textbf{\textit{Pass\_to\_T(1)}} to pass the last tool to Turtelbot\_$1$.}
        [0.3\linewidth]{\includegraphics[height=2.5cm]{s2_6.png}}
    ~
    \subcaptionbox{In the end, Turtlebot\_$1$ completes the final delivery.}
        [0.3\linewidth]{\includegraphics[height=2.5cm]{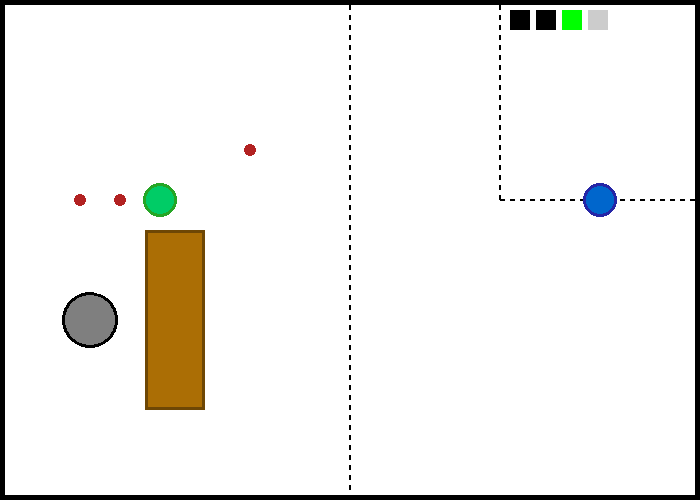}}

    \caption{Visualization of the cooperation behaviors given by the same centralized policy run in Fig.~\ref{wtds1}, but under Turtelbot moving speed $v=0.8$.}
    \label{wtds2}
\end{figure}

We notice that, in Fig.~\ref{wtds1}g, Turtlebot\_$0$ waits over there for the last tool because letting Turtelbot\_$1$ deliver the last tool will cost longer time. Then, we increase the Turtlebot's speed from 0.6 to 0.8 and would like to see how this centralized policy responds to this change. It actually outputs a new reasonable collaboration behaviors respect to this higher speed (Fig.~\ref{wtds2}). The interesting behavior changes start from Fig.~\ref{wtds2}d. 

\end{document}